\documentclass[acmtog]{acmart}
\renewcommand\footnotetextcopyrightpermission[1]{} 
\makeatletter
\renewcommand\@formatdoi[1]{\ignorespaces}
\renewcommand\footnotetextcopyrightpermission[1]{}
\makeatother

\usepackage{booktabs} 

\usepackage{xspace}
\usepackage{amsmath}
\usepackage{commath}
\usepackage{romannum}
\usepackage[ruled]{algorithm2e} 

\SetAlFnt{\small}
\SetAlCapFnt{\small}
\SetAlCapNameFnt{\small}
\SetAlCapHSkip{0pt}
\acmJournal{TOG}

\setcopyright{none}

\begin{document}
\title{Learning Aesthetic Layouts via Visual Guidance}

\author{Qingyuan Zheng}
\orcid{0000}
\affiliation{%
  \institution{University of Maryland, Baltimore County}
  \country{USA}}
\email{qing3@umbc.edu}
\author{Zhuoru Li}
\orcid{0000}
\affiliation{%
	\institution{ProjectHAT}
	\country{China}}
\email{hatsuame@gmail.com}
\author{Adam Bargteil}
\orcid{0000}
\affiliation{%
	\institution{University of Maryland, Baltimore County}
	\country{USA}}
\email{adamb@umbc.edu}

\begin{abstract}

We explore computational approaches for visual guidance to aid in creating aestheticlly pleasing art and graphic design.  Our work complements and builds on previous work that developed models for how humans look at images. Our approach comprises three steps. First, we collected a dataset of art masterpieces and labeled the visual fixations with state-of-art vision models. Second, we clustered the visual guidance templates of the art masterpieces with unsupervised learning. Third, we developed a pipeline using generative adversarial networks to learn the principles of visual guidance and that can produce aestheticlly pleasing layouts. We show that the aesthetic visual guidance principles can be learned and integrated into a high-dimensional model and can be queried by the features of graphic elements. We evaluate our approach by generating layouts on various drawings and graphic designs. Moreover, our model considers the color and structure of graphic elements when generating layouts. Consequently, we believe our tool, which generates multiple aesthetic layout options in seconds, can help artists create beautiful art and graphic designs.

\end{abstract}

%
%


%
%


\begin{teaserfigure}
	\centering
	\includegraphics[width=\linewidth]{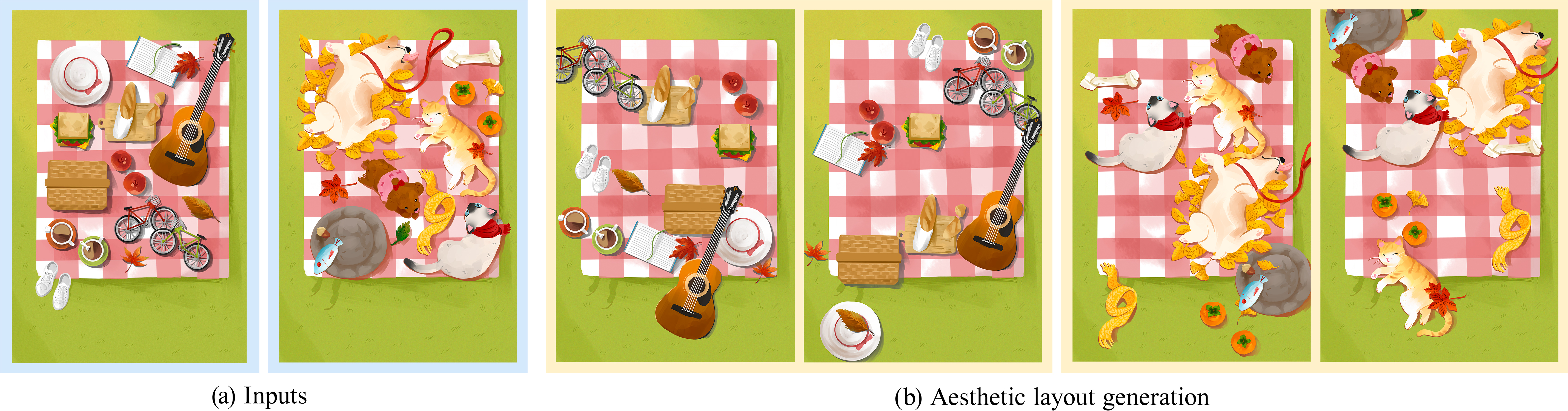}
	\caption{We propose a pipeline including a generative neural network to learn aesthetic layouts from art masterpieces. In a real-world application, our pipeline can help the artist to layout their drawing or graphic design. The artists provide their own drawings as (a). Our pipeline places the graphic elements with the help of eye fixation predictions. Then the neural network inside our pipeline generates aesthetic wireframe layouts in seconds. The artists can drag the elements to new positions suggested by our automatic pipeline to produce eye-catching visual art (b).}
	\label{fig:teaser}
\end{teaserfigure}

\maketitle
\pagestyle{plain}
\thispagestyle{empty}

\section{Introduction}

\begin{figure}[h]
	\centering
	\includegraphics[width=\linewidth]{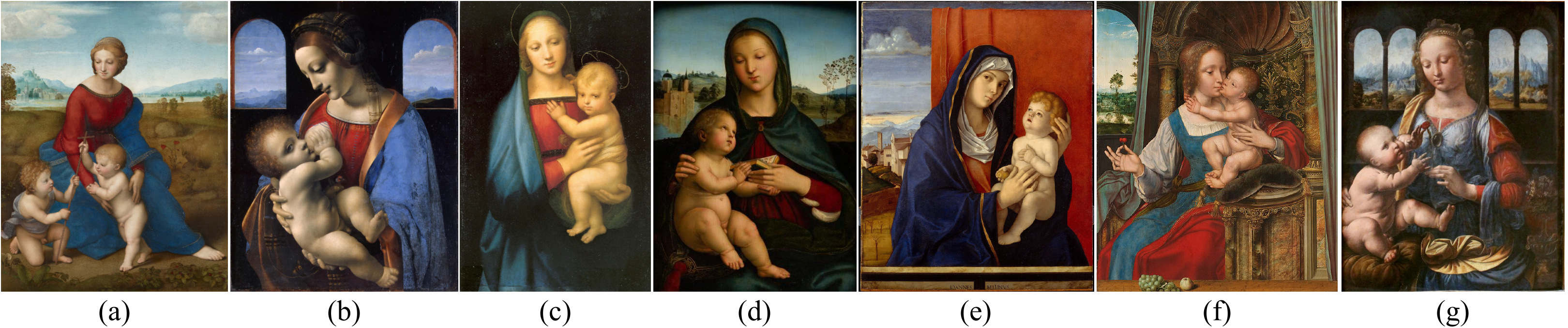}
	\caption{Madonna and Child paintings from the 14th to 16th century. (a) The three-person layout stands out among the paintings of the same subject because the triangle visual guidance implies harmony and steady state.}
	\label{fig:madonna}
\end{figure}

\begin{figure}[!ht]
	\centering
	\includegraphics[width=\linewidth]{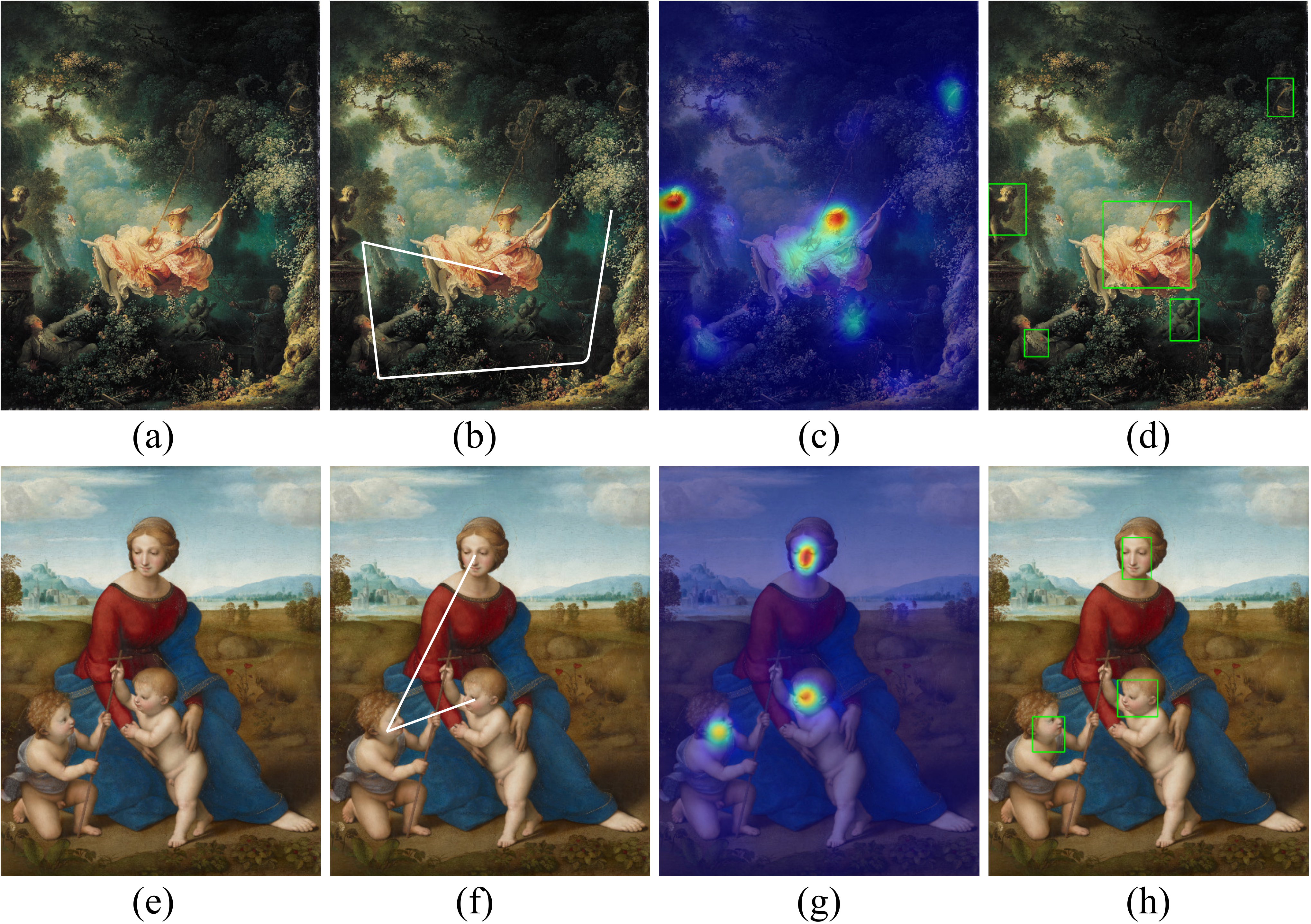}
	\caption{Examples of visual guidance in art masterpieces.  (a) [The Swing, Jean-Honor{\'e} Fragonard]. (e) [Madonna in the Meadow, Raphael]. (b) and (f): Visual guidance marked by human. (c) and (g): Visual fixation automatically predicted by vision model. (d) and (h): Interested area marked by (c) and (g).}
	\label{fig:example-layout}
\end{figure}

What causes you to like a painting at first glance? The answer could be the harmonic color or appealing content, but visual guidance plays a crucial role in eye-catching drawings. For instance, in Fig~\ref{fig:madonna}, many renowned artists depicted Madonna gently cradling the child on her lap or holding him in her arms. Why does Raphael's depiction (Fig~\ref{fig:madonna}(a)) stand out among so many paintings on the subject? Beacuase Raphael eschews the two-person layout and explores a triangle layout. The large isosceles triangle layout (Fig~\ref{fig:madonna}(a)), which is constructed by the silhouette of the three people in the foreground, makes the viewers feel stable and harmonic. Moreover, the small acute triangles of visual guidance (Fig~\ref{fig:example-layout}(f)) are more aesthetically pleasing.

Humans' aesthetic preferences regarding layout in paintings, photography, and graphic design have been studied from theoretical \cite{arnheim1965art}, empirical \cite{roberts2007mastering}, and psychological \cite{palmer2013visual} perspectives. There are many composition/layout templates, which have been summarized empirically, to make photographs and paintings more appealing. For example, common layout templates include centering and symmetry, leading lines, diagonals, horizontal, vertical, triangles, ``S''/``C''/``X''/radial shape, centered by light, and the golden ratio. Appealing layouts have been summarized and written in books \cite{photolayout, roberts2007mastering}. Even so, not every beginner quickly becomes a master. Our goal is to help beginning artists along the learning curve and serve as an assistant for masters. Our pipeline can suggest a variety of layouts that point out new positions for graphic elements to enhance the creative experience.

In this paper, we will explore visual guidance in art masterpieces and how to make aesthetically pleasing art in three steps:
\begin{enumerate}
\item Collect a dataset of well-known art masterpieces.
\item Explore an end-to-end application that can automatically output new aesthetic layouts from the given original drawing, graphic design, or photograph.
\item Analyze and explore the rules of layouts through graph kernels and unsupervised clustering.
\end{enumerate}

For readers not familiar with fine art, we will elaborate on some of the terminology we use before describing our technical approach in detail. ``Layout'' usually refers to the high-level structure of a drawing, for example, the vertical, horizontal, diagonal, and ``S'' shape layout. In our paper, we will use ``layout'' in a broad sense; for example, ``different layout'' means the elements in the drawing have different positions. ``Visual guidance'' usually means the visual fixations and visual path in a drawing which guide the audience to sense the drawing at first glance, for example, the white line in Figure~\ref{fig:example-layout}(b) and (f). We will follow the narrow definition of fine art in this paper. Layouts have been empirically summarized into fixed templates well enough by professional artists, similar but not the same as visual guidance. Visual guidance has a number of principles including ``make the center of mass (visual fixations) on the center of canvas'' and ``do not make the fixations/contrast/color regions/visual path length evenly distributed on the canvas'', which cannot be formed to fixed templates. Therefore, we will explore how to summarize great visual guidance in computational methods and encapsulate the professional design of visual guidance into a high-dimensional end-to-end model to help the artist's drawing.

The aesthetic-aware layout generation proposed by this paper potentially can be solved by another approach -- graph matching. In such an approach, there will be an interactive application; when the artist provides a drawing, the graph encoded visual guidance will be matched to the dataset and the program will output a score, representing how well the input drawing matches the art masterpieces dataset. So the artist can adjust the layout interactively in order to achieve a high aesthetic score. However, this solution must have a large amount of data and must have dozens or hundreds of people to score each piece of data, in a process similar to O'Donovan and colleagues on color compatibility\cite{odonovan2011}. We cannot assert that every art masterpiece is worth ``5 stars'' or that other drawings are ``0 stars''. We think treating aesthetics as an adversarial problem is questionable. Therefore, without having a large amount of annotated data, we choose not to score the aesthetic of a drawing.

In this paper, we explore the hypothesis that visual guidance templates, which are hidden in beautiful eye-catching drawings, can be extracted, clustered, and integrated into a high-dimensional model with computational approaches. After collecting the eye-catching art masterpieces, we use a state-of-art visual fixation model to mimic eye tracking and obtain the visual guidance data in Section~\ref{section:Dataset}. We encode visual guidance hyper information into graphs and form graph kernels to solve the unsupervised graph clustering problem -- summarize dozens of visual guidance templates through hundreds of art masterpieces in Section~\ref{section:explore-templates}. Independently, we develop a pipeline to output new aesthetic-aware layouts for the input drawing with a feature encoder, generative network, two adversarial networks, and a differentiable renderer in Section~\ref{section:gan-method}. We evaluate our aesthetic-aware layout generation model on various graphic design styles and drawings and conduct a user study with the help of six professional artists. Some of the shown evaluation results are made by the artists. 

\section{Related Work}
\label{section:related-work}
We first review the literature on art and design in computer vision and graphics and then review relevant work in machine learning.

\subsection{Art and Design in Computer Vision and Graphics}

Jahanian and colleagues~\shortcite{jahanian2015learning} gave support to Arnheim's visual balance theory by analyzing a large photography dataset. The authors fit the mixture of Gaussian to the saliency maps of the photography dataset and then obtained the images' hot spots. They showed that the hot spots aligned with Arnheim's theory of visual balance. Zheng and colleagues~\shortcite{zheng2019content} proposed a content-aware layout generation method. The authors created their dataset by collecting magazine pages from several different common categories, annotating the elements of each page based on semantics, and extracting keywords from the text. They also used a GAN-based pipeline. The resulting network takes as input a couple of discrete images indicating the magazine's content, keywords, and attributes. The output is a series of potential layouts with the labels of text, image, and headline regions. Lu and colleagues~\shortcite{lu2020exploring} explored visual information flows in infographics. They collected a large dataset of infographics, extracted the visual elements, and automatically connected the visual elements based on Gestalt principles. Then they used PCA and DBSCAN to cluster the dataset to generate 12 kinds of storytelling patterns in infographics. Moreover, they discussed the design protocols of visual information flow based on the summarized design patterns. O'Donovan and colleagues~\shortcite{o2014learning} proposed an energy-based model that can synthesize a layout from a small set of examples. They optimize an existing layout by designing objective functions that encode design principles such as alignment, symmetry, and white space. Unfortunately, the optimization is slow. The authors extended this work and provide interactive speeds for layout suggestions \cite{o2015designscape}. In contrast to content-aware layout generation \cite{zheng2019content} and layout optimization \cite{o2014learning, o2015designscape}, we will explore aesthetic-aware layouts.
Additionally, content-aware layout generation \cite{zheng2019content} requires as input keywords and categories; in contrast, our work must have a differentiable renderer to convert the plain layout to an RGB image. These differences make any direct comparison difficult. While comparisons with work on layout design~\cite{o2014learning, o2015designscape} might be informative, no open-source codes exist for this work.

Moving to the topic of art and design in computer vision and graphics, Reddy and colleagues~\shortcite{reddy2020discovering} proposed a differentiable compositing function for optimizing the design of pattern structure. The authors also demonstrate that image pyramids with differentiable compositing improves the gradients in optimization. They brightly add the image pyramid by comparing the different sizes of images in the loss function. We will borrow the idea from this paper to implement a differentiable compositing function in our work. Adding an image pyramid into a deep neural network to improve the gradient flow is a common trick when the expected images lack rich information; for example, the edge detection of Liu and colleagues~\shortcite{liu2017richer} would output the image from an intermediate layer in the deep network and then compare with ground truth.

Regarding photography, there is previous work using a saliency map to auto crop photographs \cite{8259308} to center the main object. The color compatibility and aesthetics in graphic design have also been studied in computational methods with large annotated datasets \cite{odonovan2011, odonovan:2014:cfcolor} to help find harmonic and appealing color combination in a drawing. Regarding indoor scenes planning, Graph2Plan \cite{hu2020graph2plan} works on transferring the user-provided layout graphs into a floor plan, GRAINS \cite{li2019grains} helps generate various plausible indoor scenes by learning from a large dataset of hierarchical scene structures and PlanIT \cite{wang2019planit} synthesizes realistic indoor scenes constructed by selective 3D models, with the relation graph gained from scene input. LayoutGMN \cite{patil_cvpr21} proposed a graph matching neural network which can compute the similarity scores between two structural layouts, for example, the floor plan.

\subsection{Generative Networks}

A generative model, such as Deep Brief Net \cite{hinton2006fast}, is an unsupervised learning method under the umbrella of machine learning.  Generative models can learn to synthesize images. While the preceding work in generative models have the shortcomings of high computational cost and training difficulty, VAE (Variational Autoencoder) \cite{kingma2013auto} and GAN (Generative Adversarial Networks) \cite{NIPS2014_5ca3e9b1} partly show the prospect of training stably. To improve stability, training process, and generalization to different tasks, conditional GAN, Wasserstein GAN, least squares GAN, CycleGAN, Pix2Pix, StarGAN, sinGAN, etc. \cite{mirza2014conditional, arjovsky2017wasserstein, mao2017least, zhu2017unpaired, isola2017image, choi2018stargan, shaham2019singan} advance the GAN study in image synthesis and content creation. GAN is expandable to have multiple generative or adversarial networks like CycleGAN \cite{zhu2017unpaired}, and D2GAN \cite{NIPS2017_6860}. LayoutGAN \cite{li2018layoutgan} proposed a wireframe-based approach to optimize randomly placed 2D elements to match the target layout. While LayoutGAN focuses more on generative adversarial networks -- treating pixels as layout to generate images, this work has different priors and tasks from ours.

\begin{figure*}[ht]
	\centering
	\includegraphics[width=\linewidth]{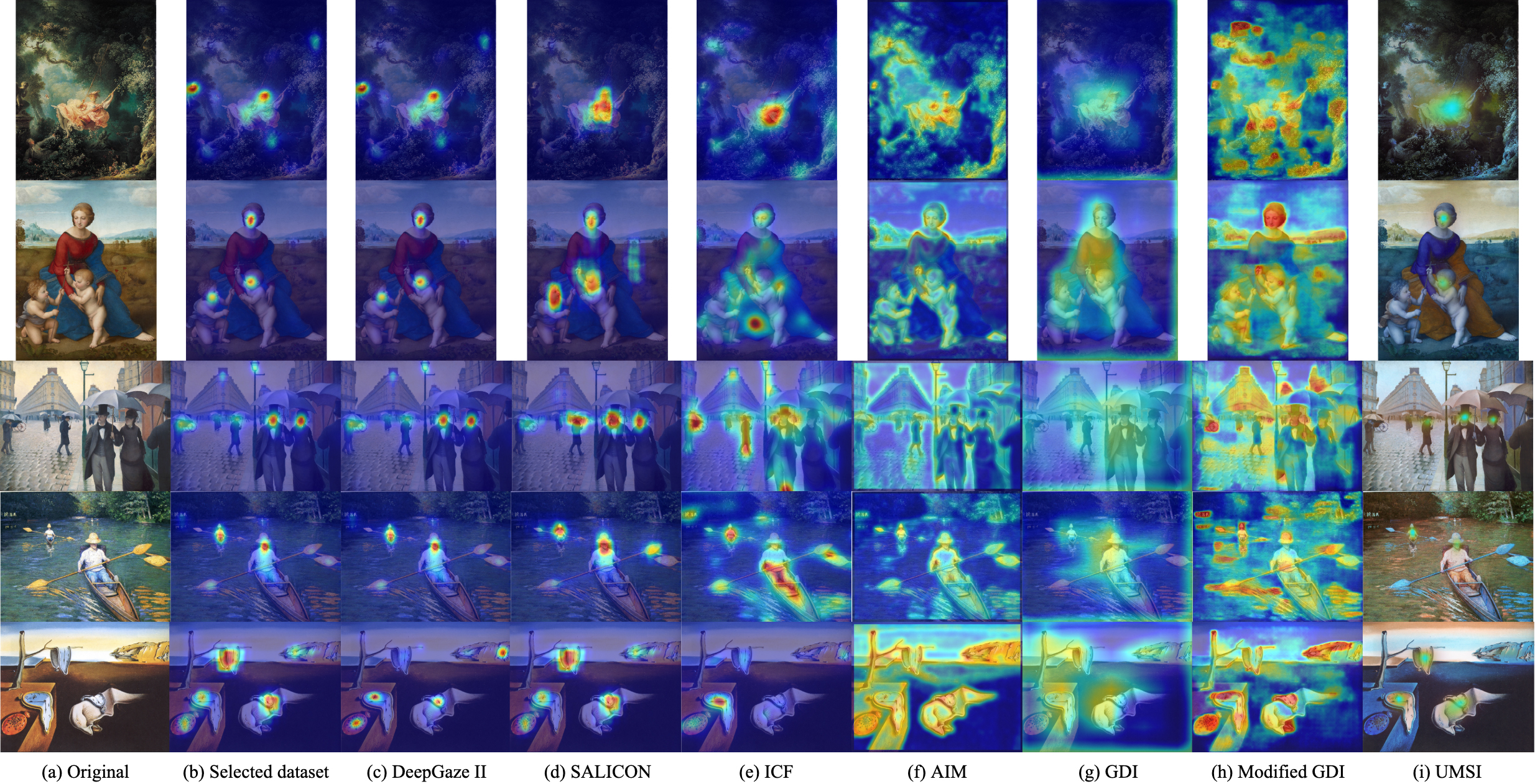}
	\caption{We produce paired dataset from original paintings (a) by selecting the eye fixation resulted by the state-of-art saliency models. We select our datasets (b) from the predictions of DeepGaze II \cite{kummerer2016deepgaze} (c) and SALICON \cite{jiang2015salicon} (d). We also show the candidates resulted from ICF \cite{Kummerer_2017_ICCV} (e), AIM \cite{bruce2006saliency} (f), GDI \cite{o2014learning, bylinskii2017learning} (g), a modified model trained with GDI dataset by us (h), and UMSI \cite{fosco2020predicting} (i).}
	\label{fig:dataset-example}
\end{figure*}

\section{Dataset}
\label{section:Dataset}
In this section, we describe the criteria of how we collect the art masterpieces, generate the visual saliency maps, and construct the layout scene graphs. We discuss why we only generate visual saliency but forgo the visual path information.

\subsection{Data Collection}
We collected well-known masterpieces from WikiArt \cite{wikiart} and the Art Institute of Chicago \cite{chicagomuseum} throughout the art movements of Early and High Renaissance, Baroque, Rococo, Romanticism, Impressionism, Post-Impressionism, Modern Art, Contemporary Art, Abstract Art, Chinese Art, Japanese Art, etc. The masterpieces are in the shape of paintings in a 2D domain. Sculptures are not included. The media of selected paintings are mainly canvas, oil, and watercolor.

We manually select the famous artworks of which layout was one of the contributions. An eye-catching painting is not only comprised of its layout; content, colors, shading, and stokes also could be the main contribution. Thus, we ignore the artworks with trivial layout. Many of the artworks in our dataset are collected from the art movements such as Renaissance, Rococo, Impressionism, and Post-Impressionism, where the layout was vital. We selected a few paintings that are highly abstract but still imply a novel visual layout.

To ensure our dataset has high quality, we collected only a few of the most well-known artworks from each famous artist, based on the ``most-featured'' ranking in WikiArt and art history knowledge. Because even master artists also have many mediocre artworks, we collected 5 to 15 paintings from any given artist. Even for artists such as Raphael, Van Gogh, Monet, or Dali we did not collect more than 15 paintings from their work. Because we want to learn the layout from genuine eye-catching paintings, we have to left out the mediocre works.
We also kept the sourced genres and styles as varied as possible. 
After data processing, we manually delete the artworks where automatic visual saliency prediction fails. Our final data set consists of 550 paintings.  

\subsection{Data Processing}
We use current visual fixation models to mimic eye tracking to obtain visual guidance data.
Alternatively, we could use eye-tracking equipment 
to collect visual saliency and gaze path.
Such eye-tracking equipment captures the visual hierarchy, but compiling this data from multiple subects remains a difficult open problem. Current automatic fixation models work well enough for our puposes. Visual hierarchy consists of eye fixation spots and the visual path connecting the fixation. Multiple people's visual path of one drawing have too many variations to be merged.
Interstingly, a psychological study by Urano and colleagues~\shortcite{uranogood} found that eye-tracking data of good graphic design has similar visual hierarchies among the subjects. In contrast, poor graphic design has a miscellaneous visual hierarchy. Given that the visual hierarchy of multiple subjects is highly correlated to good graphic design, we choose to use current automatic models to predict the visual hierarchy. However, our experiments with STAR-FC \cite{wloka2018active} on our painting dataset yeilded poor results. Consequently, we do not include visual path in our dataset because neither eye-tracking equipment, nor saliency models produce accurate and clean data.

We feed the paintings in our dataset Fig~\ref{fig:dataset-example}(a) to current visual saliency models. Figure~\ref{fig:dataset-example} shows the results of the visual saliency models -- DeepGaze II \cite{kummerer2016deepgaze} (c), SALICON \cite{jiang2015salicon} (d), ICF \cite{Kummerer_2017_ICCV} (e), AIM \cite{bruce2006saliency} (f), GDI \cite{o2014learning, bylinskii2017learning} (g), a modified model trained with GDI dataset by us (h) and UMSI \cite{fosco2020predicting} (i). 

All of the shown heat maps in Fig~\ref{fig:dataset-example}(c)-(i) are the candidates that we use to select eye fixation predictions for our dataset. UMSI has its own particular post-processing; we show the heat map resulted from the original implementation of UMSI. Notably, GDI and UMSI are trained on graphic design data with annotated visual importance. The other works are trained on natural images. Presumably, GDI and UMSI are a good fit for our data -- art masterpieces. However, GDI and UMSI are not a good fit for our purpose. We prefer a heat map that can be segmented well since we want to analyze the layout, and we assume the viewers sees the paintings for a short time. So we manually review and select only one visual saliency map from (c) to (i) with the standard that (1) it can be segmented well, and (2) it properly marks the visual importance region. Fig~\ref{fig:dataset-example}(b) shows the selected data from (c) to (i). Finally, we have paired data (a) original painting and (b) its visual saliency map for future use.

\begin{figure*}[ht]
	\centering
	\includegraphics[width=\linewidth]{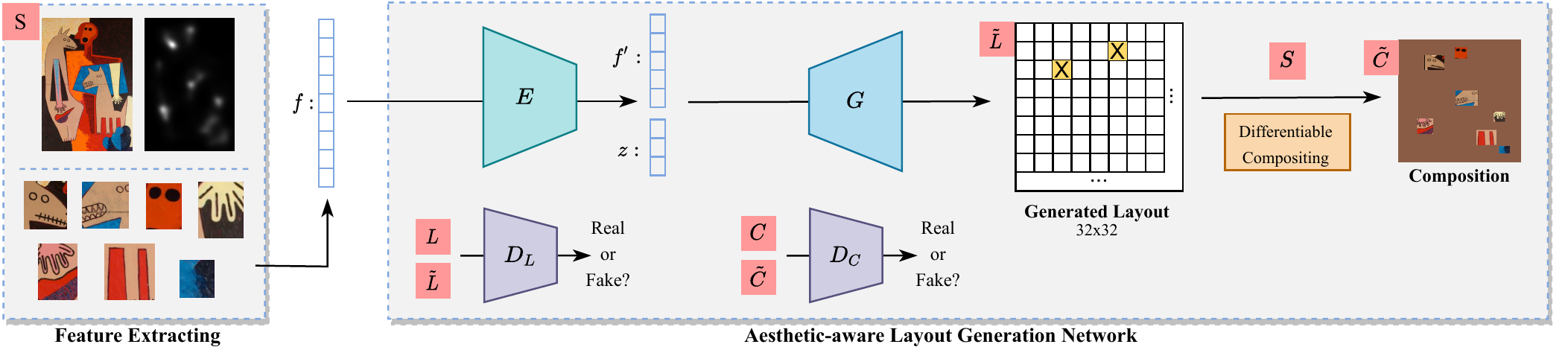}
	\caption{The pipeline of our aesthetic layout generation system. We extract and encode the regions of interests of a drawing into a feature vector $f'$ (section~\ref{section:encoder}). The generative network $G$ generate wireframe layout conditioned on the $f'$ (section~\ref{section:gd}). We input the number of graphic elements into the bottleneck of $G$. We composite the image patches in set $S$ to match the new layout $L$ using a differentiable renderer (section~\ref{section:layout}). $D_{L}$ and $D_{C}$ discriminate the real/fake wireframe layouts and real/fake composited images respectively. The encoder $E$, generator $G$ and two discriminators $D_{L}$ and $D_{C}$ are trained jointly.}
	\label{fig:pipeline}
\end{figure*}

After we produce the visual saliency map paired with art masterpieces, we consider how to construct the graphs to represent the layout of the art masterpieces and how to make use of the dataset to create aesthetic-aware layout generation in Section~\ref{section:explore-templates} and Section~\ref{section:gan-method}.

\section{Generative Aesthetic-aware Layouts}
\label{section:gan-method}
Our goal is to build a pipeline that can learn and integrate aesthetic-aware layout principles into a non-linear function. This well-trained non-linear function has learned numerous aesthetic layouts from various drawings in a large latent space; hence, we can treat the features of input drawings as a key to query the latent space to find an appropriate aesthetic-aware layout. 

\subsection{Feature Encoder}
\label{section:encoder}
For each art masterpiece, the eye fixation areas are segmented by the Wastershed algorithm, which results in $n$ local patches $\{I_{1}^{k}, ..., I_{n}^{k}\}$ from the original $k$ number of drawings. For example, in Figure~\ref{fig:graph}, local patches circled by green boxes are $I$. In our case, the specific value of $ n \in \{1, ..., N\}, N=13 $ may be different for different drawings. After segmenting the images, we obtain the feature vectors $\{a_{1}^{k}, ..., a_{n}^{k}\}$ for each $I_{n}^{k}$ local patch by passing the $I_{n}^{k}$ to a pre-trained ResNet34 \cite{he2016deep} and output the flattened feature before the last max pooling layer. Then the input of the feature encoder is the sum of $a_{n}$:
\begin{equation}
f^{k} = \sum_{n=1}^{N}  a_{n}^{k}.
\label{eq:03}
\end{equation}

In the training process, we input a batch of $ f^{k} $ in $\mathbb{R}^{b \times 512} $ into the feature encoder, where $b$ is the batch size. Batch training is one of the reasons we sum up the patches' features of each art masterpiece. Then the $ f^{k} $ are encoded to a smaller size $ f'^{k} $ in $ \mathbb{R}^{b \times 128}$ as the input of Generative network $G$. In our early research, we tried to use Graph Convolutional Networks \cite{kipf2017semi} with an adjacency matrix to aggregate the patches' features; but, in our experiemnts, the positions of elements congregate to the center of canvas; hence we shifted to our method that simply adds up the features.

\subsection{Generative and Discriminator Networks}
\label{section:gd}
We follow the Wasserstein GANs (WGAN) with gradient penalty \cite{arjovsky2017wasserstein, NIPS2017_892c3b1c} to build our generative network and two discriminator networks. The generative network will partly be conditioned on the $ f' $ resulting from the encoder. We call the concatenation of $ f' $ and $z$ as $z'$. Therefore the generative network does not wholly generate random layouts that align with the data distribution from our dataset but generate the layouts conditioned on the appearance of input.

When humans see a scene, only the eye fixation area is clear; the remaining areas are blurred. Vision research follows this rule to build algorithms of eye fixation sequence prediction \cite{wloka2018active}. We also follow this rule to mimic human perception in our layout generation model. Therefore,
we only include the eye fixation areas when creating features for training our model of aesthetic layout.
Furthermore, we use the average color of the areas in which humans are not interested, as the background both in our ground truth and the $\tilde C$ images in differentiable compositing shown in Figure~\ref{fig:pipeline}.

The input of the generative network is $z'$. The output of the generative network is a $32\times 32$ wireframe layout named $\tilde L$. The points in $\tilde L$ represent the center positions of the graphic elements. We use the information of discrete elements in $S$, which are the patches in the original art masterpiece, and $ \tilde L$ to composite an RGB image $\tilde C$. The ground truth in our dataset is denoted as $L$ and $C$ for wireframe layout and composited images respectively. The generative loss is

\begin{equation}
\mathcal{L}_{G} = - \lambda_{1} \underset{\tilde L \sim \mathbb{P}_{g}}{\mathbb{E}} \big[ D\,(\tilde L) \big] - \
     \lambda_{2} \underset{\tilde C \sim \mathbb{P}_{g}}{\mathbb{E}} \big [ D\,(\tilde C) \big ]. 
\label{eq:gloss}
\end{equation}
 
 The discriminator $D_{L}$ for the wireframe layout has a similar architecture as the Wasserstein GANs.
 The discriminator loss for $D_{L}$ is
 
 \begin{equation}
 \mathcal{L}_{D_{L}} = \lambda_{3} \big ( \underset{\tilde L \sim \mathbb{P}_{g}}{\mathbb{E}} \big[ D\,(\tilde L) \big] - \
 \underset{L \sim \mathbb{P}_{r}}{\mathbb{E}} \big [ D\,(L) \big ] + \
 \lambda \underset{\hat L \sim \mathbb{P}_{\hat L}}{\mathbb{E}} \big [ \big( \, \norm{ \nabla_{\hat L} D\,(\hat L) }_{2} - 1 \big )^{2} \big ]. \big)
 \label{eq:dloss1}
 \end{equation}

 \begin{figure*}[tb]
	\centering
	\includegraphics[width=\linewidth]{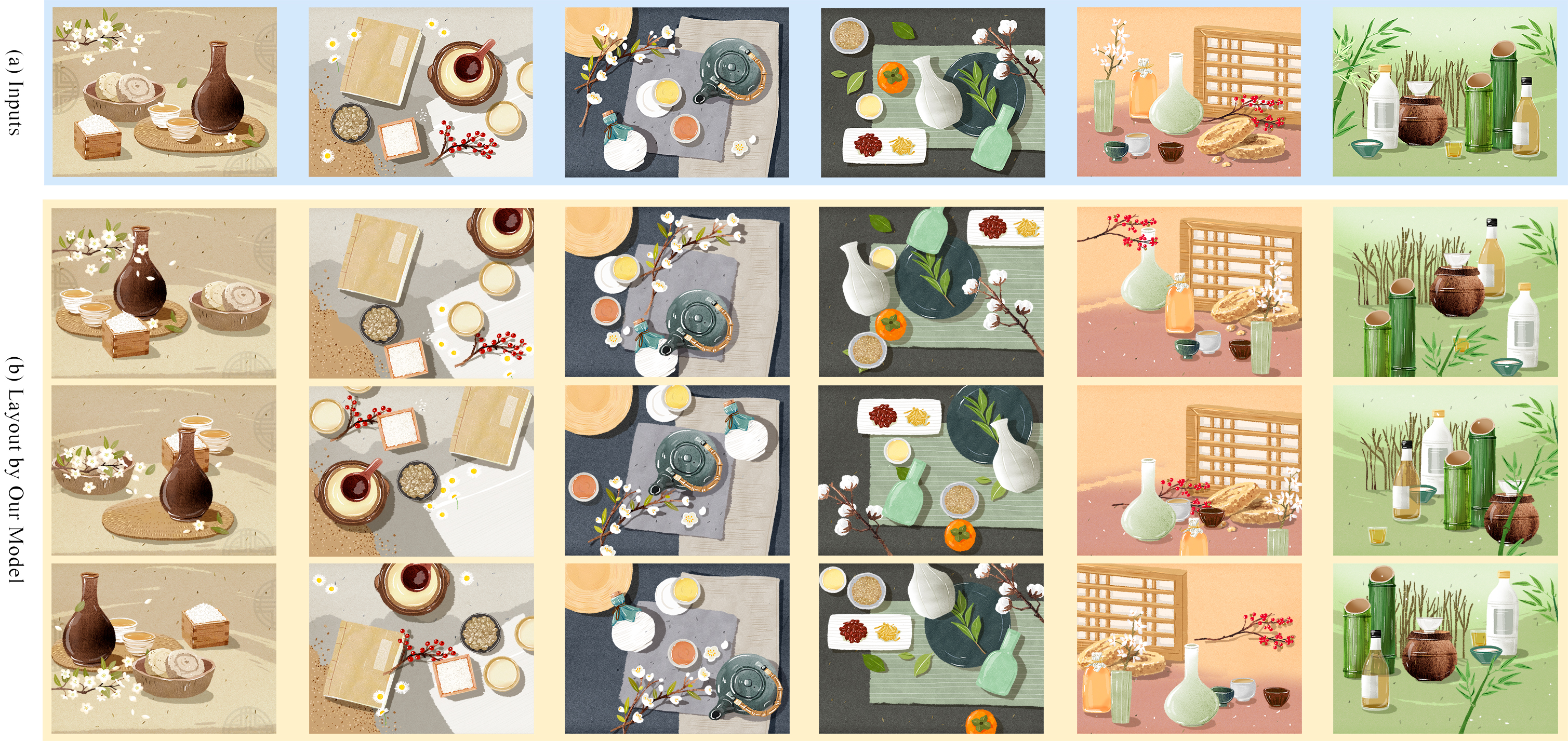}
	\caption{Evaluations on cartoon graphic design with 5 to 10 elements. (a) Original graphic design. (b) Layout again using our model. We observe our model layout the elements with the same hue or same features together in majority generative evaluations, such as checking the tiny white flowers.}
	\label{fig:minsu}
\end{figure*} 
 
  $\mathbb{P}_{r}$ and $\mathbb{P}_{g}$ denote the sampling from the real data and generative data distribution. While $\mathbb{P}_{\hat L}$ represents sampling between the $\mathbb{P}_{r}$ and $\mathbb{P}_{g}$. WGAN encourages the discriminator to give different scores for the real and fake data; hence the gradients will be as large as possible, which is reflected in the third part in eq(\ref{eq:dloss1}).
 
 The second discriminator $D_{C}$ which takes as input the composited RGB image has the same discriminator loss as $D_{L}$'s. 
 
 \begin{equation}
\mathcal{L}_{D_{C}} = \lambda_{4} \big ( \underset{\tilde C \sim \mathbb{P}_{g}}{\mathbb{E}} \big[ D\,(\tilde C) \big] - \
\underset{C \sim \mathbb{P}_{r}}{\mathbb{E}} \big [ D\,(C) \big ] + \
\lambda \underset{\hat C \sim \mathbb{P}_{\hat C}}{\mathbb{E}} \big [ \big( \, \norm{ \nabla_{\hat C} D\,(\hat C) }_{2} - 1 \big )^{2} \big ]. \big )
\label{eq:dloss2}
\end{equation}

In order to improve the differentiability and gradient flow of images $\tilde C$, we add image pyramids to the discriminator $D_{C}$ with the help of the open-source differentiable computer vision library Korina \cite{eriba2019kornia}. Let us define the downsampling function of Kornia as $f_{n}^{k}(\cdot)$ and the downsampling convolution in $D_{C}$ as $f_{n}^{d}(\cdot)$. Also, we use $f_{n}^{k}$ and $f_{n}^{d}$ to represent the result of the current function. In the $(n+1)$-th stage of the discriminator, we do the operation as

\begin{equation}
f_{n+1}^{d} \big( \, [f_{n}^{k}, f_{n}^{d} ] \, \big), \quad 
f_{n+1}^{k} \big( f_{n}^{k} \big).
\label{eq:dsample}
\end{equation}
When $n=1$, the operation is $f_{1}^{k}\,(C)$ and $f_{1}^{d} \, (C)$. Note the $f^{d}(\cdot)$ here contains the weights that can be back propagated and updated, while $f^{k}(\cdot)$ is just down sampling the image $C$ again and again differentiably. At the last stage, the $f^{k}$ and $f^{d}$ are concatenated together then flattened to be the input of a linear layer and then the network finally computes the validity. The coefficients are set to: $\lambda = 10$, $\lambda_{1}=\lambda_{2}=\lambda_{3}=\lambda_{4}=0.2$.

\subsection{Layout Representation}
\label{section:layout}
We use one anchor to find the exact position of a graphic element. We use a discrete $32\times32$ wireframe, where each anchor is a non-zero positive number, representing the center of a graphic element. The scale factors of each graphic element in the new canvas will be recorded independently. In addition, there is a crucial information -- the order we should follow to place the graphic elements in set $S$ to the wireframe $\tilde L$ shown in Figure~\ref{fig:pipeline}. For our ground truth, we set the values of the wireframe based on both the overlap with a graphic element and the importance of the element (measured by area). We sort importance and use the function $\exp(-0.1 \times i)$, $i \in \mathbb{N}_{0}$, where $i$ is the integer ranking of importance. For wireframe nodes that do not overlap graphic elements, the value will be $0$ in layout $L$. In both training and evaluation modes, we take the first $n$ largest values in the wireframe layout resulting from the generative network to be the new anchors.

\begin{figure*}[ht]
	\centering
	\includegraphics[width=\linewidth]{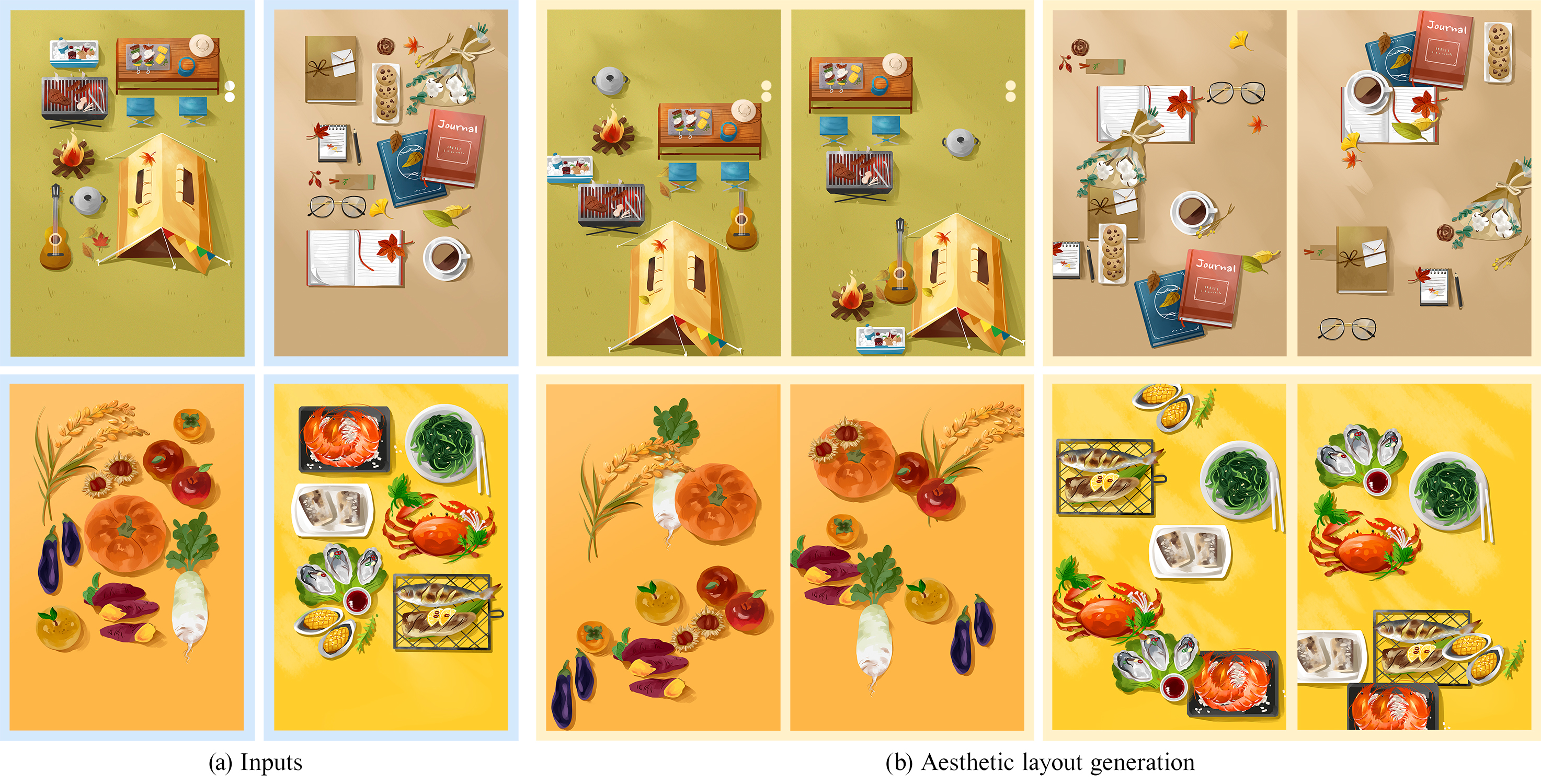}
	\caption{Evaluations on cartoon graphic design with 7 to 14 elements. (a) Original graphic design. (b) Layout the original drawing again using our model.}
	\label{fig:show1}
\end{figure*}

From our experiments and user study, our $32 \times 32$ wireframe is fine-grained enough to determine the centers of graphic elements for 2K and 4K drawings (e.g. the pictures in figure~\ref{fig:teaser} and~\ref{fig:show1} are made in 4K then resized to fit in figures.). In our early research, we used a list of floating numbers, for example, "element width / canvas width", to represent the exact positions of the graphic elements. The predicted new positions congregated into the central areas of the canvas because of loss functions like mean square error. Additionally, learning a regression to precise positions that are floating numbers involves too large a search space.

In order to avoid complex regression, we use the wireframe layout $L$ which is a $32 \times 32$ matrix in our pipeline. Later, we will create a RGB image $C$ as shown in Figure\ref{fig:pipeline} with differentiable compositing modified from \cite{reddy2020discovering}. Besides improving the gradient, we believe the composited RGB image can avoid unnecessary overlapping graphic elements when training the networks.

To compute our final images, first the coordinates in the new canvas are computed to find the RGB values from the graphic elements in their local coordinates. Then, we use bilinear interpolation to sample the graphic elements into the new canvas. In this step, the discrete graphic elements are transformed to the same size canvas in multiple layers. Then we soft merge the multiple layers using the alpha channel of the graphic element; hence the overlapping regions of graphic elements are merged softly with weights that respect opacity. Hereafter, the discrete graphic elements are transformed to the same canvas {\em differentiably}. Notably, bilinear sampling improves the differentiability, but it is not smooth/differentiable everywhere; hence we add some image pyramid operations in the discriminator $D_{C}$ to further improve the differentiability. The gradient and differentiability need to be treated carefully in this work, otherwise, the lack of the image pyramid makes the pipeline fail to converge.

\subsection{Implementation Details}
The feature encoder has three linear layers with Leaky ReLU as activation functions. The generative network has two stages of Linear-Batch Normalization-Leaky ReLU; in the bottleneck, we reshape the vector to square image size. We concatenate the anchor numbers, representing how many elements in one drawing, into the bottleneck of the generative network. Then we have two stages of 2D transpose convolution-Batch Norm-Leaky ReLU. At last, there is a 2D transpose convolution with Tanh as an activation function. Following the WGAN, we do not have any batch normalization in the two discriminator networks. The general architecture of the two discriminator networks are similar, with three stages of convolution 2D-Leaky ReLU; in the bottleneck, we reshape the square size tensors to vectors, then perform Linear-Leaky ReLU-Linear to result in a validity number. We initialize the weights of the generative network and two discriminator networks with normal distribution values. 

We trained our network with Adam optimizer setting the beta1~$=0.5$, beta2~$=0.999$ and learning rate~$=0.0002$ for $400$ epochs. The generative network will be updated only once by the original WGAN after the discriminator network gets updated every five times. Unlike WGAN, we update the generative and discriminator network with the same frequency. 

\begin{figure*}[th!]
	\centering
	\includegraphics[width=\linewidth]{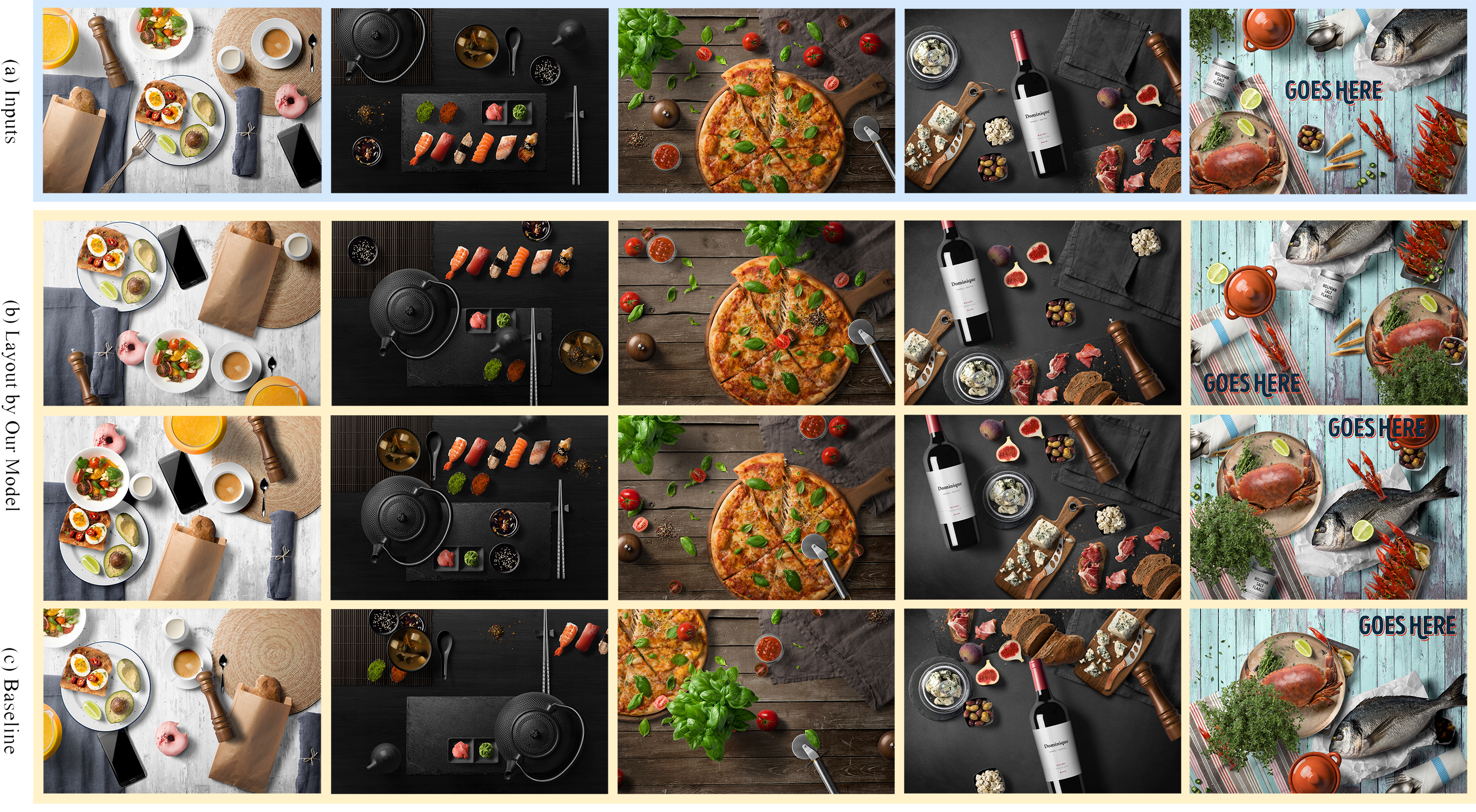}
	\caption{Evaluations on photograph. (a) Original photograph. (b) Layout again using our model. (c) Baseline using uniform noise to random a layout.}
	\label{fig:show2}
\end{figure*}

\section{Results and Application}
\label{section:result}
We discuss and show the results from the algorithm in section~\ref{section:gan-method} here. Note the content of section~\ref{section:gan-method} and section~\ref{section:explore-templates} are independent algorithms and the results of the two sections do not affect each other. 

We do not compare with any related work \cite{o2014learning, o2015designscape, zheng2019content}. The detailed reasons are stated above (section~\ref{section:related-work}). Generally, the related work targets UI/poster graphic design with a few text and image regions, while we target digital art/graphic design with graphic elements around a dozen; most prior work focuses on reasonable layout and avoids overlapping elements; we focus on aesthetic layout and encourage pleasing overlapping. We create two baselines for ourselves. In the first baseline, we output the layout results from our untrained networks. In the second baseline, we generate uniform noise, then select the first $n$ large numbers to form the layouts. We aim to compare our results with random layouts using the two baselines. When evaluating the baselines, the authors and recruited professional artists agree that the baselines' results are too random and meaningless, so we did not spend more time making complete images for baselines. We show the results of the second baseline in Figure~\ref{fig:show2} and~\ref{fig:10result}. The positions in the first baseline are evenly distributed because our untrained networks are initialized with normal noise.

\textbf{Application:} Since we have the eye fixation maps and segmentation as described in section~\ref{section:gan-method}, the graphic elements in the original drawing will be marked with color boxes; with the ranking of box areas in descending order, the boxes in the original drawing will be colored from one fixed color palette. On the other hand, we color the positions of first $n$ large numbers from the wireframe layout in descending order from that color palette too. When laying out the graphic elements, the center of the color box should overlap with the dot position, marked in the same color. Hence, the paired color boxes and points can guide the users to drag the marked graphic element to the marked position, which is in the same color. More pictures of raw output and the marks on drawings can be found in the appendix.

\begin{figure}[t!]
	\centering
	\includegraphics[width=0.9\linewidth]{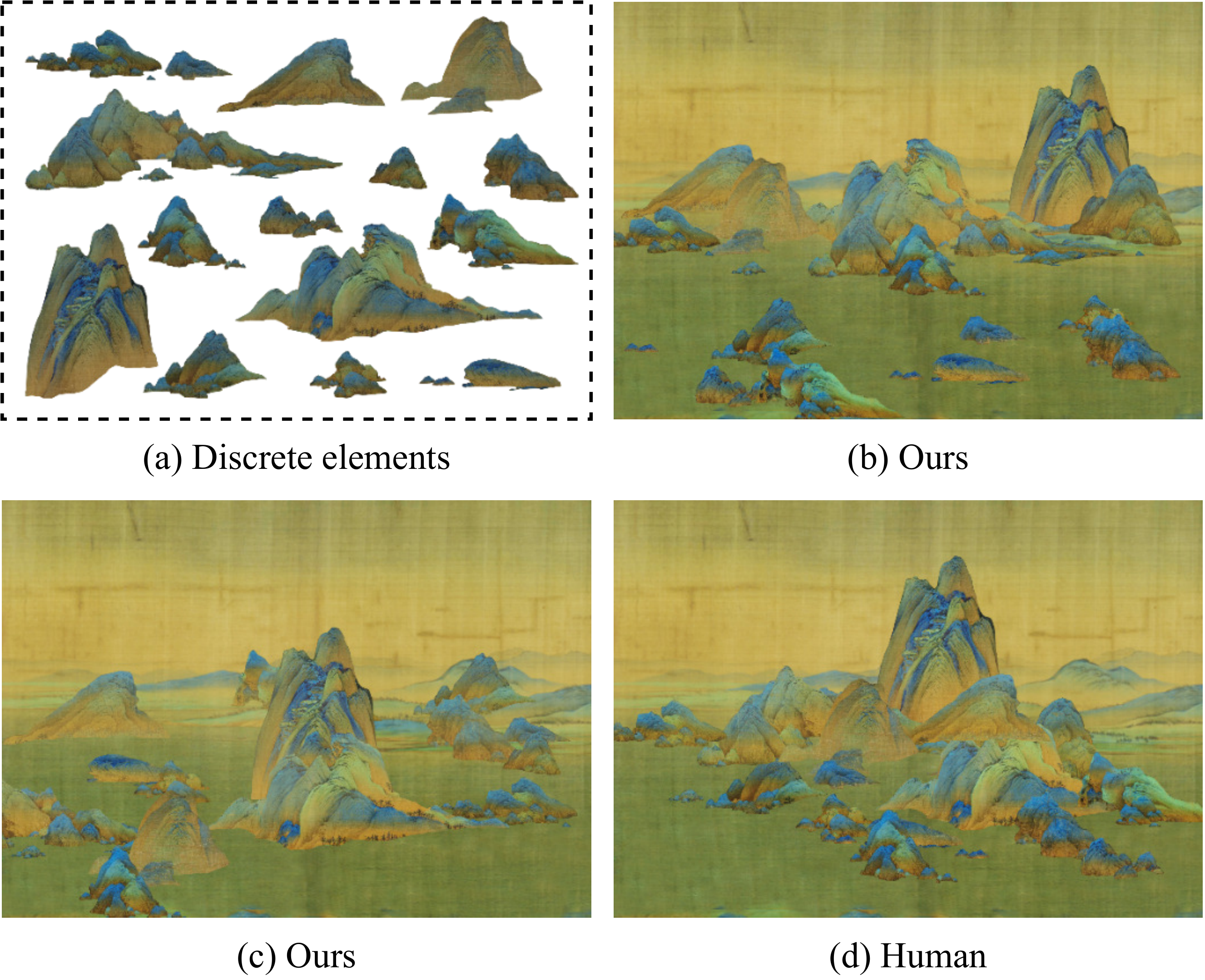}
	\caption{Evaluations on one part of \textit{A Panorama of Rivers and Mountains}, Wang Ximeng (1096-1119). This style and similar appearance rarely appear in our dataset. (a) Discrete elements from original painting as inputs. (b)-(c) Layout result using our model. (d) Layout result made by professional artist.}
	\label{fig:show3}
\end{figure}

Since the input of our generative network consists of one feature vector and one noise vector, whenever the noise vector changes, our network will output a different layout for the same drawing. Moreover, the pre-trained weights of our generative network only have 30 megabytes; therefore, we can output layouts in real-time for the artist until the output layout matching their preference.

The layout of our pipeline is not limited to the size or ratio of drawings. First, the actual input to our network is the feature of eye fixation regions. Technically, the pipeline can work on drawings in any reasonable size. Second, though the wireframe layout is $32\times32$, it can be rescaled to match the size of the drawing. We visualize the raw wireframe layout in the appendix and enlarge the dot 4$\times$ bigger for better viewing. In our user study and from our observation, the artists and we agree that the wireframe layouts are fine-grained enough and can be matched to any reasonable size/ratio of drawing/graphic design.

\begin{figure}[t]
	\centering
	\includegraphics[width=\linewidth]{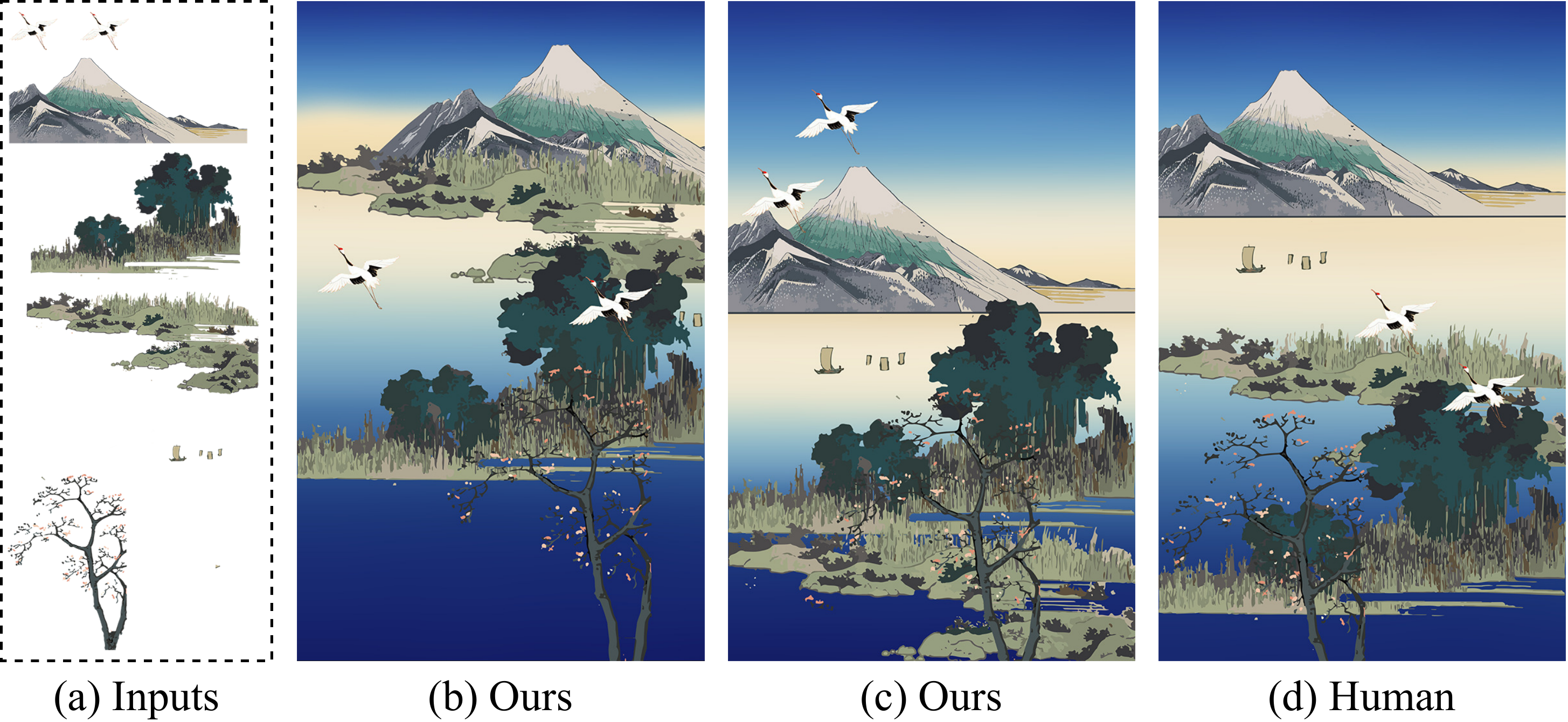}
	\caption{Evaluations on Ukiyoe. This style rarely appears in our dataset. (a) Discrete graphic elements as input. (b)-(c) Layout made by our model. (d) Layout made by human.}
	\label{fig:show5}
\end{figure}

\begin{figure}[t!]
	\centering
	\includegraphics[width=\linewidth]{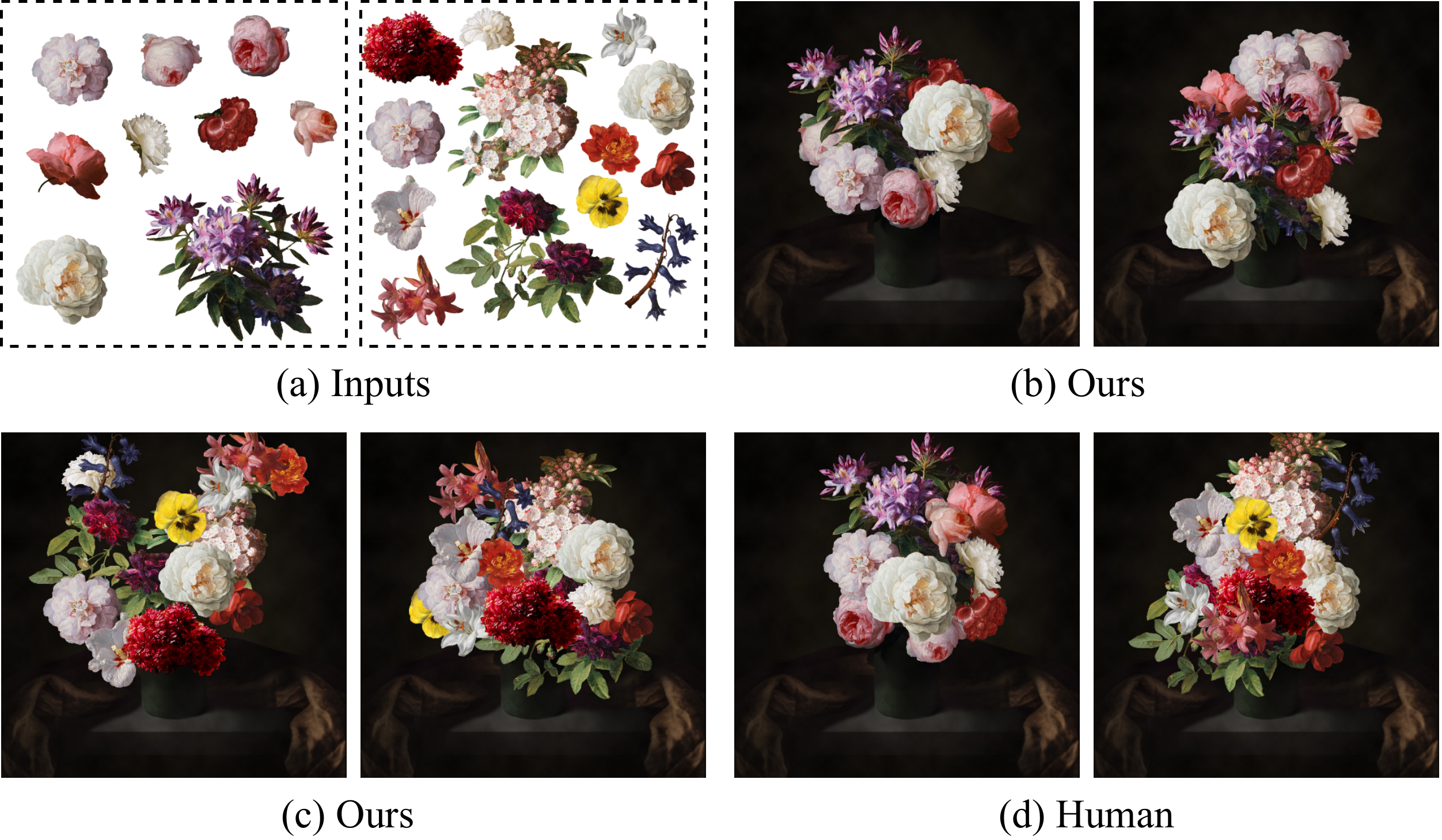}
	\caption{Evaluations on still life. (a) Discrete graphic elements as inputs. (b)-(c) Layout results made by our model. (d) Layout made by human.}
	\label{fig:show6}
\end{figure}

\textbf{Results of aesthetic layouts generation:} We show the evaluation results of our model in Figure~\ref{fig:minsu},~\ref{fig:show1},~\ref{fig:show2},~\ref{fig:show3},~\ref{fig:show5} and ~\ref{fig:show6}.

In our evaluation, we output ten layouts for one drawing/graphic design at a time. For the majority of cases, we can find at least two pleasing layouts from the results. When dragging the graphic elements in the original layered drawing file, we may have minor offsets for 0 to 2 elements, which means the center of these elements will not precisely overlap with the marks on wireframe layout because some elements cannot overlap in the physics world. Refer to the appendix for the visualization of offsets and more raw outputs.

\section{User Study}
\label{section:user}
We recruited six professional artists, three of them are photographers. First, we let the users select 5 drawings/graphic designs, which they want to work on from our evaluation set. Second, like the application section, we let users have ten layout options for each drawing. If the users are not satisfied with the ten options, we will let users generate another ten options. In our user study, none of the users required a second set of layouts, they found at least 2 pleasing results from the first round. Then we let the artists follow the procedure described in the application section, identical to how we produce the final image.
Finally, we interviewed the artists with a few questions and let them provide their feedback. We show the results that some of the artists made in Figure~\ref{fig:show2}. We summarize the positive and negative feedback from the six artists together since they have many similar feedback.

\textbf{Positive feedback}: The layouts made by our system are reasonable and plausible. Though the layouts are not as regular as the original drawing, the users think the layouts are attractive and pleasing. The users are supervised that the system can output some layout structures which human artists can do. For example, they noticed there are triangle and ``S'' structures, structures that a few small elements surrounding a more prominent element, and attractive structure even with complementary colors at the same time. The users are also satisfied with the color harmony for the whole drawing made from our system. One user noticed that some layout results in the ten options reach the principle that the center of mass for eye fixations overlaps with the center of the canvas.

\textbf{Negative feedbacks}: The system does not provide the information of rescaling and rotation factors for the graphic elements and the information of the overlapping order. It seems the system does not have a great sense of depth to account for perspective. Not every one of the generated layouts is reasonable. For some photograph layouts that users are pleased with, if they look closely, they believe some of the layouts will be recognized as computer generated because of the arrangement of the awkward arrangements of some local elements.

To sum up, all users are pleased with our system. They think it is convenient to select the preferred layouts from the ten options by previewing the hard composition (shown in the appendix) of regions of interest. All users agree that the two baselines are too random. Since we do not have a friendly user interface and detailed user instructions, two users were confused about selecting and deselecting the elements to make them able/unable to be considered in the automatic layout system. They can do this and change the regions of interest simply by retouching the visual fixation map.

\section{Discussion}
\label{section:discussion}
Given that we make the hypothesis that visual guidance in art masterpieces has guidelines and templates (e.g. the triangle template in Figure~\ref{fig:example-layout}(f)), in this section, we justify the motivation and data are meaningful for our aesthetic layout generation (section~\ref{section:gan-method}) by exploring the templates in visual guidance with graph kernels and unsupervised clustering.

\subsection{Explore the Templates in Visual Guidance}
\label{section:explore-templates}
\begin{figure}
	\centering
	\includegraphics[width=\linewidth]{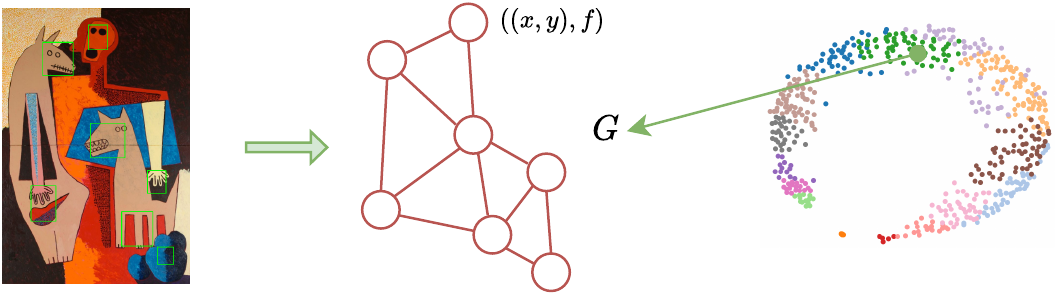}
	\caption{A schematic diagram of section~\ref{section:explore-templates}. We encode the visual guidance structures into graphs and perform unsupervised graph clustering by graph kernels to explore the regular templates of visual guidance in aesthetic-aware layouts.}
	\label{fig:graph}
\end{figure}

First of all, the templates are graphs that are not in euclidean space; accordingly, we could not use any unsupervised clustering method in euclidean space, such as k-means. Figure~\ref{fig:graph} is a sketch map describing how we construct the graph out of an art masterpiece. Since we have the drawing and the corresponding visual fixation map (in Section~\ref{section:Dataset}), we use the Watershed algorithm to segment the visual fixation map. The content circled by green boxes in Figure~\ref{fig:graph} are $\{I_{1}^{k}, ..., I_{n}^{k}\}$. After extracting the feature vectors by pre-trained ResNet34, we have the features of local patches denoted as $a^{k}:  \{a_{1}^{k}, ..., a_{n}^{k}\}$, $a^{k}\in\mathbb{R}^{n\times 512}$. In the meanwhile, the centers of the green boxes will be recorded as $p^{k}:  \{p_{1}^{k}, ..., p_{n}^{k}\}$, $p^{k}\in\mathbb{R}^{n\times 2}$. So the nodes $V^{k}=(p^{k}, a^{k})$. Furthermore, we form the weighted edges $E$ by the Delaunay triangulation algorithm. Then we use the weighted edges to form the graphs $G=(V, E)$.

We experimented on two potential solutions in the topology and graph domain -- persistent homology and graph kernel. The pre-pilot experiment shows that persistent homology is much less promising to solve this algorithm than graph kernel. The points $V$ in one piece of our dataset are very sparse -- there are at most 13 points in one drawing; thus, the representations of persistent diagrams are not distinguishable enough among different point sets. On the other hand, the raw data described in Section~\ref{section:Dataset} lack the visual path which can be used directly to form the edges for graph $G$. We explained why we do not have the visual path in Section~\ref{section:Dataset}. Hereafter, we utilize the Delaunay triangulation forming the edges to bring more neighbor information with the trade-off bringing some noise. Because another version of this algorithm without considering any edges (empty graphs) does not work, we have to automatically add the edges into $G$. 

The Pseudocode version of our algorithm is in Algorithm~\ref{alg:clustering}. After we construct the graphs $G=(V,E)$, we employ the Wasserstein Weisfeiler-Lehman (WL) graph kernels \cite{Togninalli19} to compute one kernel matrix out of all the graphs $G^{k}$. Since we have continuous attributes $a_{n}^{k}(v) \in \mathbb{R}^{m}$, ($v \in V$), the node embeddings will be computed as 

\begin{equation}
a^{h+1}(v) = \dfrac{1}{2} \: \bigg(  a^{h}(v) + \dfrac{1}{deg(v)} \sum_{u\in \mathcal{N}(v)} w \big( (v,u) \big) \cdot a^{h}(u)  \bigg),
\label{eq:01}
\end{equation}

where $u$ is the neighborhood of $v$, $h$ is the number of WL iterations, and $w$ is the edge weight between $u$ and $v$. So, for the WL-based graph embedding scheme, the WL features at iteration $h$ are as
\begin{equation}
X_{G}^{h} = [ a^{h}(v_{1}), ..., a^{h}(v_{nG}) ]^{T},
\label{eq:02}
\end{equation}

\begin{figure*}
	\centering
	\includegraphics[width=\linewidth]{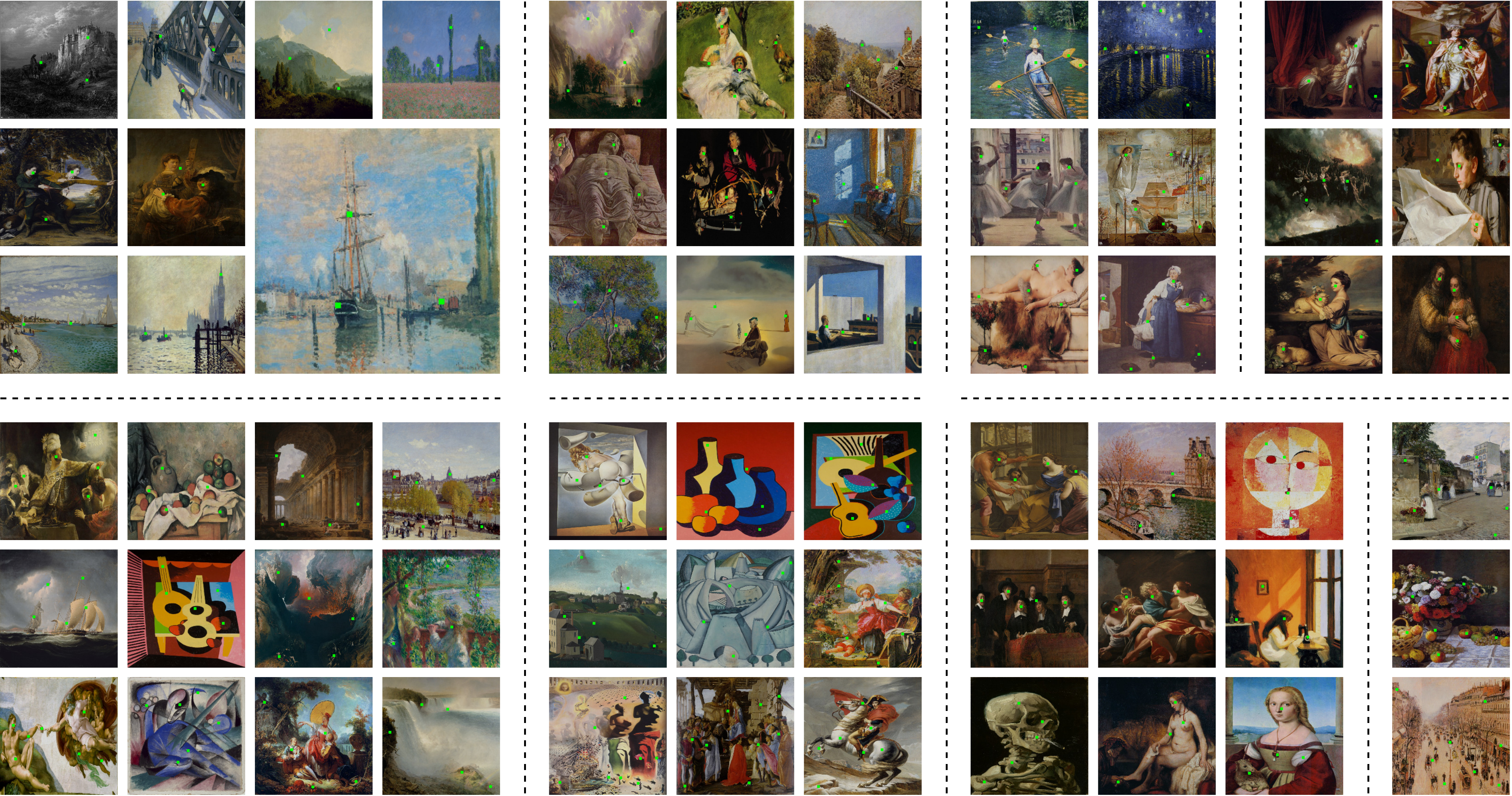}
	\caption{The unsupervised graph clustering results of section~\ref{section:explore-templates}. We overlap the original art masterpiece with the graph nodes visualized by green dots. We show a limited number of graphs from each clustering. Details of clustering results are discussed in section~\ref{section:result} (part \Romannum{1})}
	\label{fig:clustering1}
\end{figure*}

\begin{figure*}
	\centering
	\includegraphics[width=\linewidth]{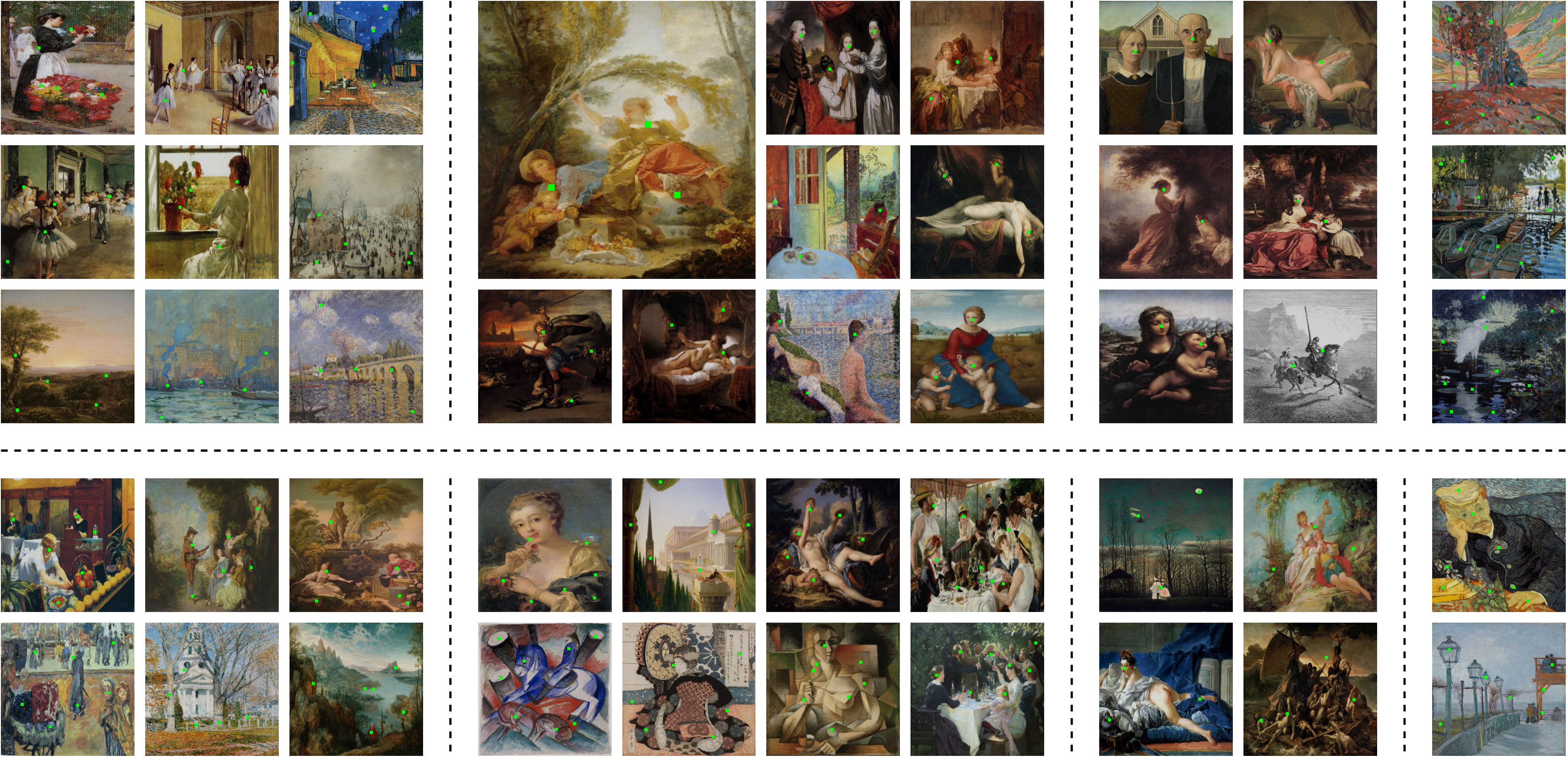}
	\caption{The unsupervised graph clustering results of section~\ref{section:explore-templates}. (part \Romannum{2})}
	\label{fig:clustering2}
\end{figure*}

where $X_{G}^{h} \in \mathbb{R}^{nG \times m}$ are the node features of graph $G$ at iteration $h$, the $i$-th row of $X_{G}$ is the node embedding of $v_{i}$. Then the node embeddings of graph $G$ at iteration $h$ are the concatenation of $(X_{G}^{0}, ..., X_{G}^{H}), h \in \{0, ..., H\}$. In our case, the initial node attributes/features could either be the features of local patches $ a_{n}^{k} $ or the center coordinates of local patches $ p_{n}^{k} $. Here we choose $ a_{n}^{k} $ because of more favorable results by experiments. At last, the pairwise graph kernel values between graphs are computed by the ground distance $d: \mathbb{R}^{m} \times \mathbb{R}^{m} \rightarrow \mathbb{R} $ which is euclidean distance between each pair of nodes and then the Wasserstein distance. Once we have the kernel matrix of the graphs, we can perform the classical unsupervised clustering algorithms using the kernel matrix as precomputed similarity matrix. 

\begin{algorithm}
	\SetAlgoNoLine
	\KwIn{$k$ Art masterpieces $A^{k}$ and the cooresponding eye fixation map $F^{k}$.}
	\KwOut{$m$ clusterings of art masterpieces. In each clustering, $A^{k}$ have similar visual guidance.}
	\For{each art masterpiece $A$ and eye fixation map $F$
	}{
		$ROI$ (regions of interest) = WaterShed($F$)\;
		$I^{k}$ = local patches in $A$ found by $ROI$\;
		$a^{k}$ = features from ResNet($I^{k}$)\;
		$p^{k}$ = centers of $I^{k}$\;
		$p^{k}$ = uniform $p^{k}$ to $32 \times 32$ grid\;
		$faces$ = Delaunay($p^{k}$)\;
	}
	$Adjacency$ = WeightedAdjacencyFromFaces($faces$, $p$)\;
	$G$ = constructed by $Adjacency$\;
	$G^{k}$.node["feature"] = $a^{k}$\;
	$kernel\,matrix$ = Wasserstein-Weisfeiler-Lehman-Graph-Kernel($G$)\;
	Hierarchical clustering by precomputed $kernel\,matrix$\;
	\caption{Unsupervised Layouts Clustering}
	\label{alg:clustering}
\end{algorithm}

\subsection{Results of Graph Clustering}
We show the results of graph clustering in Figure~\ref{fig:clustering1} and~\ref{fig:clustering2}. Since the dataset consists of 550 art masterpieces and each clustering contains up to 34 pieces of drawings, we only show samples from one clustering in the figures. For about 60\% to 70\% of the art masterpieces, we observe they have recognizable templates for visual guidance, such as triangle, ``H'' shape, a large circle shape, half-circle shape, big ``S'' shape, one triangle with one fixation far away and some cannot be rephrased verbally. For the other graphs, which are clustered together but we cannot see recognizable templates, we think it is partly due to the limitation of algorithms and partly due to the perception of humans. 

The unsupervised graph clustering supports the hypothesis we made in the Introduction. There do exist templates for aesthetic visual guidance. Most likely, the templates are numerous or cannot be summarized empirically by humans, especially when artists need to consider the appearance of each element for aesthetic layouts. Therefore, it is meaningful and worthy of developing a model to generate various aesthetic layout options to help artists.

\section{Limitations and Future Work}
Since our layout generation system is not semantics-aware, the system cannot deal with situations like requiring vehicles to be at the bottom of the canvas. We have not implemented the function that lets some of the elements have fixed positions, then allows the system to automatically layout the remaining elements; additionally, the system needs to consider all the elements to figure out the aesthetic visual guidance. This function could potentially be developed by a denoising autoencoder. 

The raw output of our network, which is the new center positions of graphics elements, causes elements to overlap. Though we added differentiable compositing in our network trying to reduce this issue, our testing results ``in the wild'' show that the issue remains. Additionally, besides the positions of the graphic elements, the outputs of our network do not contain the information of scaling and rotation information, which is critical for the artists. The limitations of our dataset cause this issue since it is difficult to gather detailed data in -- ``bad'' layout design and ``good'' layout design for the same graphic elements and content.

\section{Conclusion}
We support our hypothesis that beautiful drawings/graphic design have principles of visual guidance that can be clustered and learned by computational approaches with our experiments. First, we form the graphs from 550 art masterpieces and perform unsupervised graph clustering. We find meaningful templates of visual guidance from the graph clusterings. Second, we propose a creative pipeline to learn the aesthetic layouts from art masterpieces via visual guidance. Then the well-trained model can generate various aesthetic-aware layouts to help the artists in drawings and graphic design. We showed pleasing results of laying out the original drawings again and laying out the discrete graphic elements. Thus, we support the hypothesis -- aesthetic layout principles of beautiful drawings/graphic design can be integrated into a high-dimensional model with a computational approach, and the aesthetic layouts can be queried conditioned on the appearance of the elements in drawing/graphic design.

\bibliographystyle{ACM-Reference-Format}
\bibliography{sample-bibliography}

\appendix

\begin{figure*}
	\centering
	\includegraphics[width=0.9\linewidth]{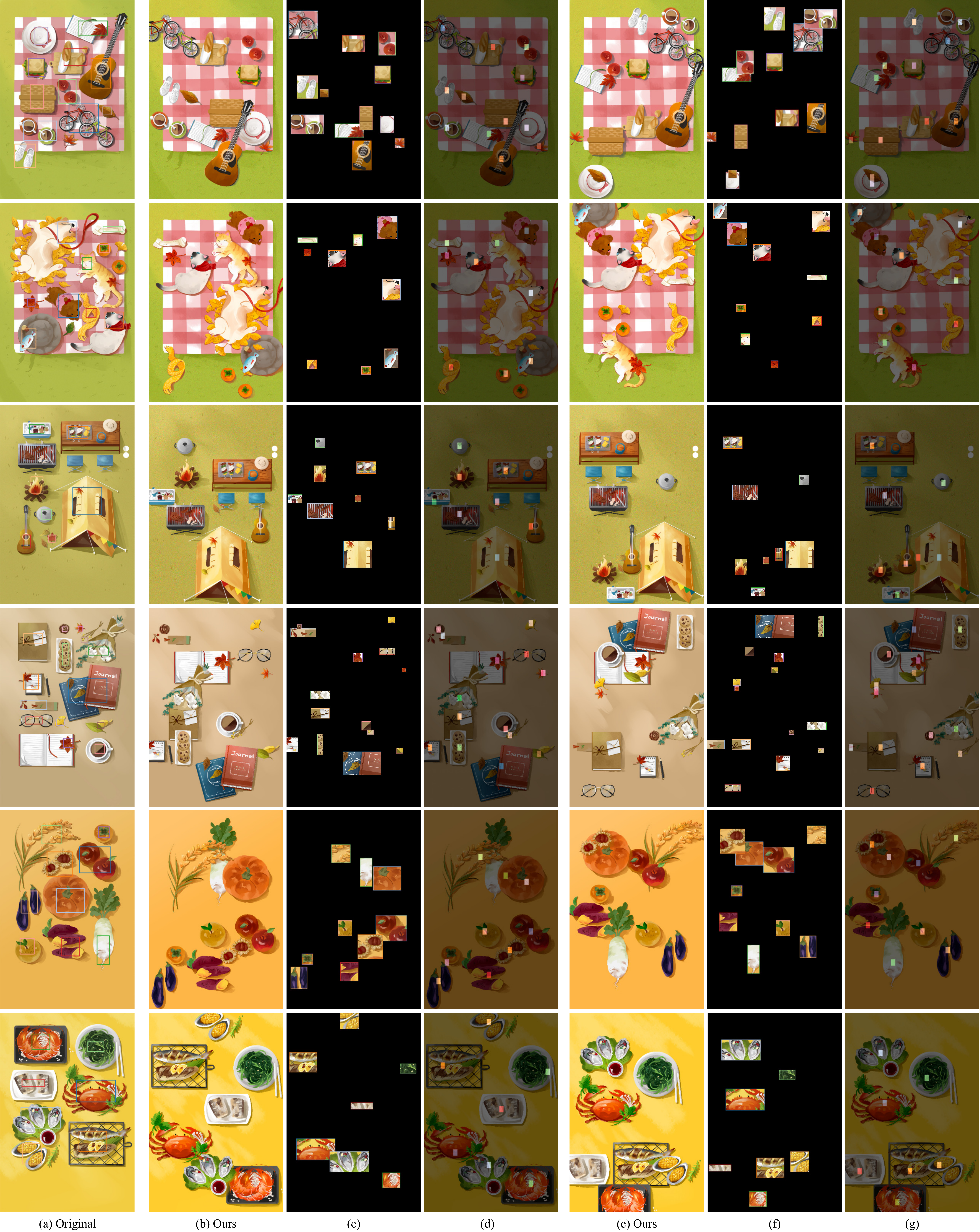}
	\caption{Supplementary materials of Figure~\ref{fig:teaser} and~\ref{fig:show1} with raw layout outputs and composited results. As we mentioned in section~\ref{section:gan-method}, the local regions of interest are marked with color boxes in the original drawings (a). When laying out the graphic elements, the center of the color box in (a) should overlap with the dot in the same color (d) (g). We strictly follow the dot positions in most cases. Minor offsets may occur in a few elements; hence we show the composition of results shown in our paper and their raw dot positions (d) (g). The dots are 4$\times$ bigger than the actual pixels in the wireframe layout output of our model.}
	\label{fig:apx1}
\end{figure*}

\begin{figure*}
	\centering
	\includegraphics[width=\linewidth]{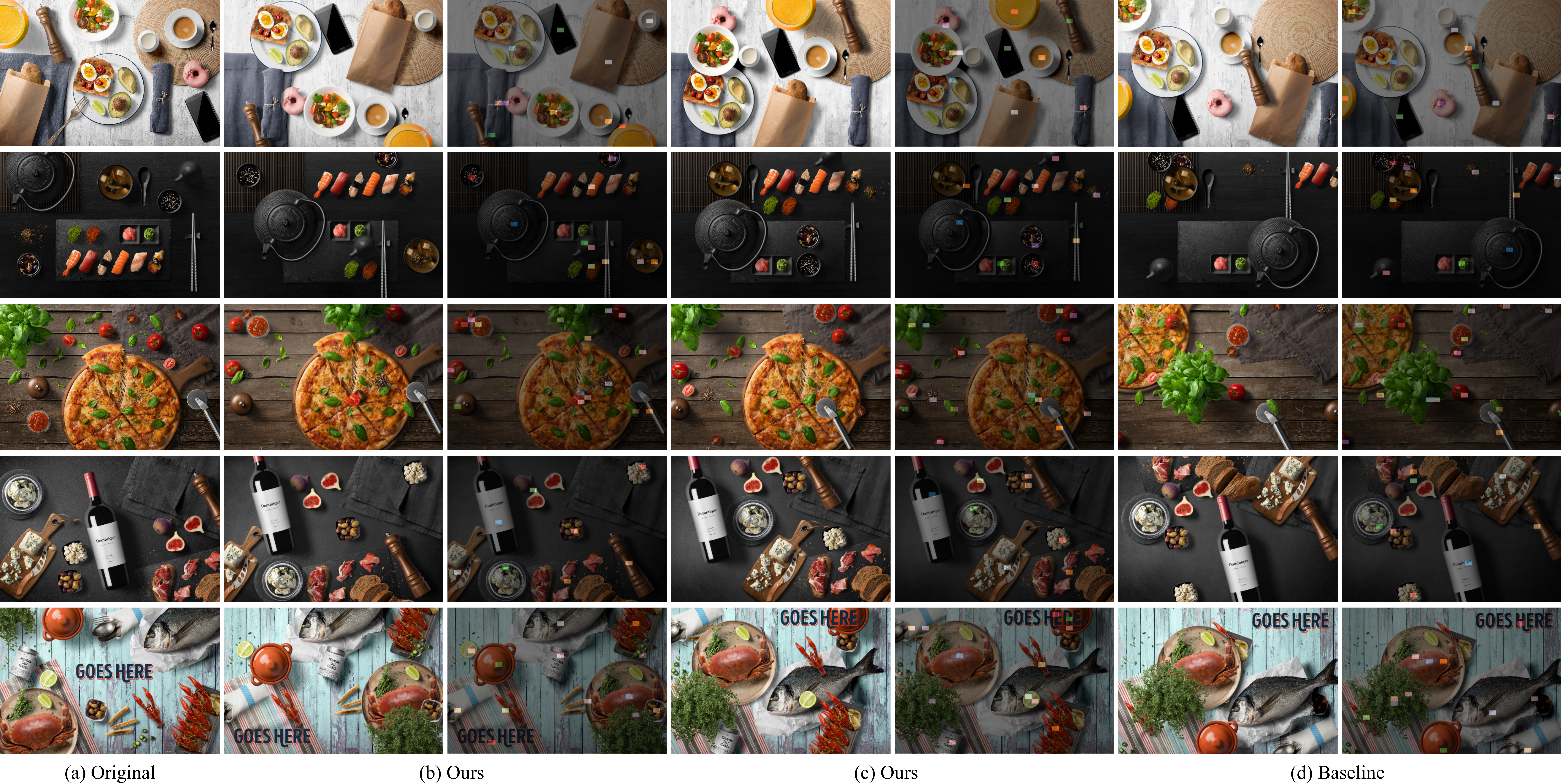}
	\caption{Supplementary materials of Figure~\ref{fig:show2}.}
	\label{fig:apx2}
\end{figure*}

\begin{figure*}
	\centering
	\includegraphics[width=0.9\linewidth]{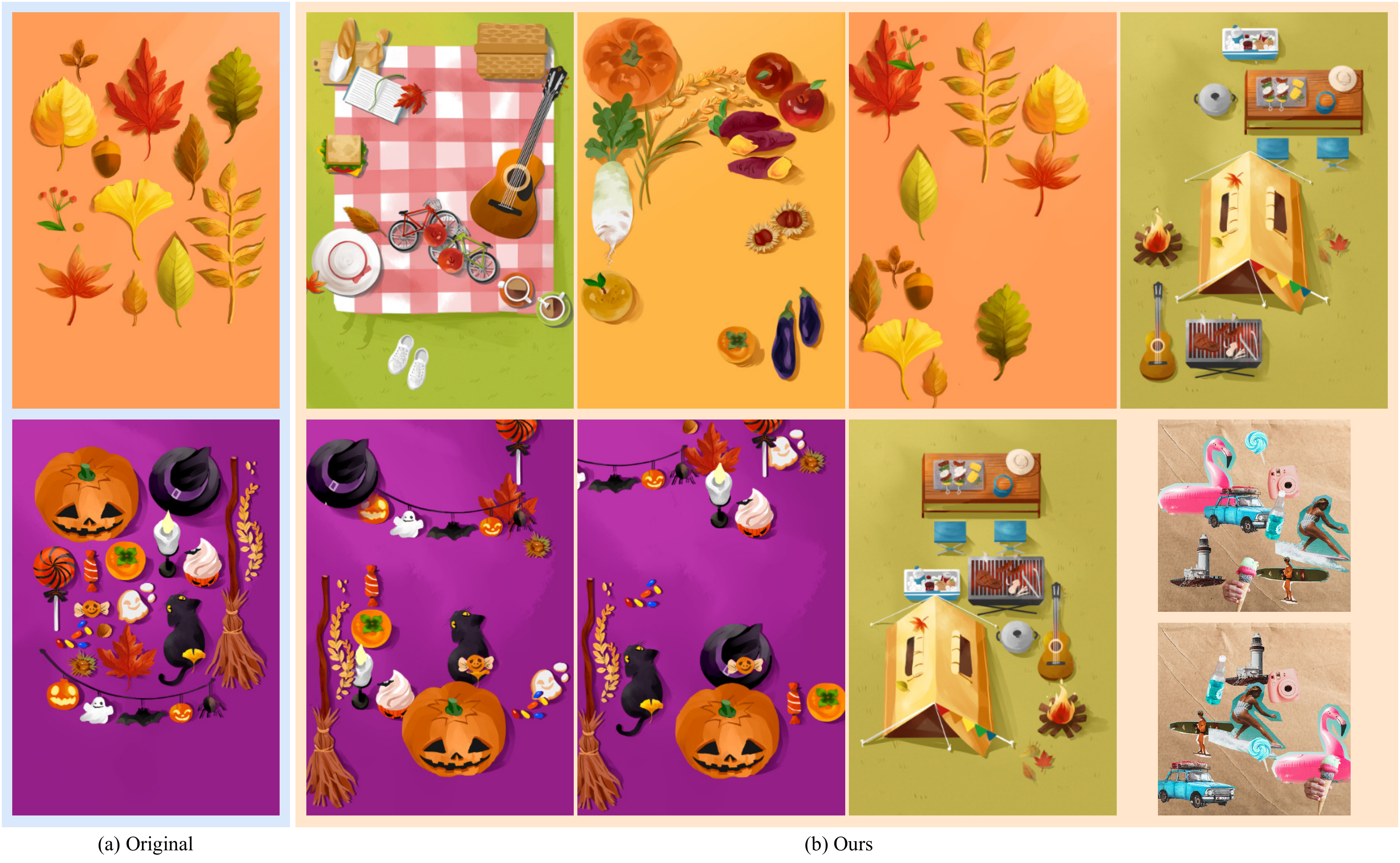}
	\caption{More results from user study (section~\ref{section:user}).}
	\label{fig:apx4}
\end{figure*}

\begin{figure*}
	\centering
	\includegraphics[width=\linewidth]{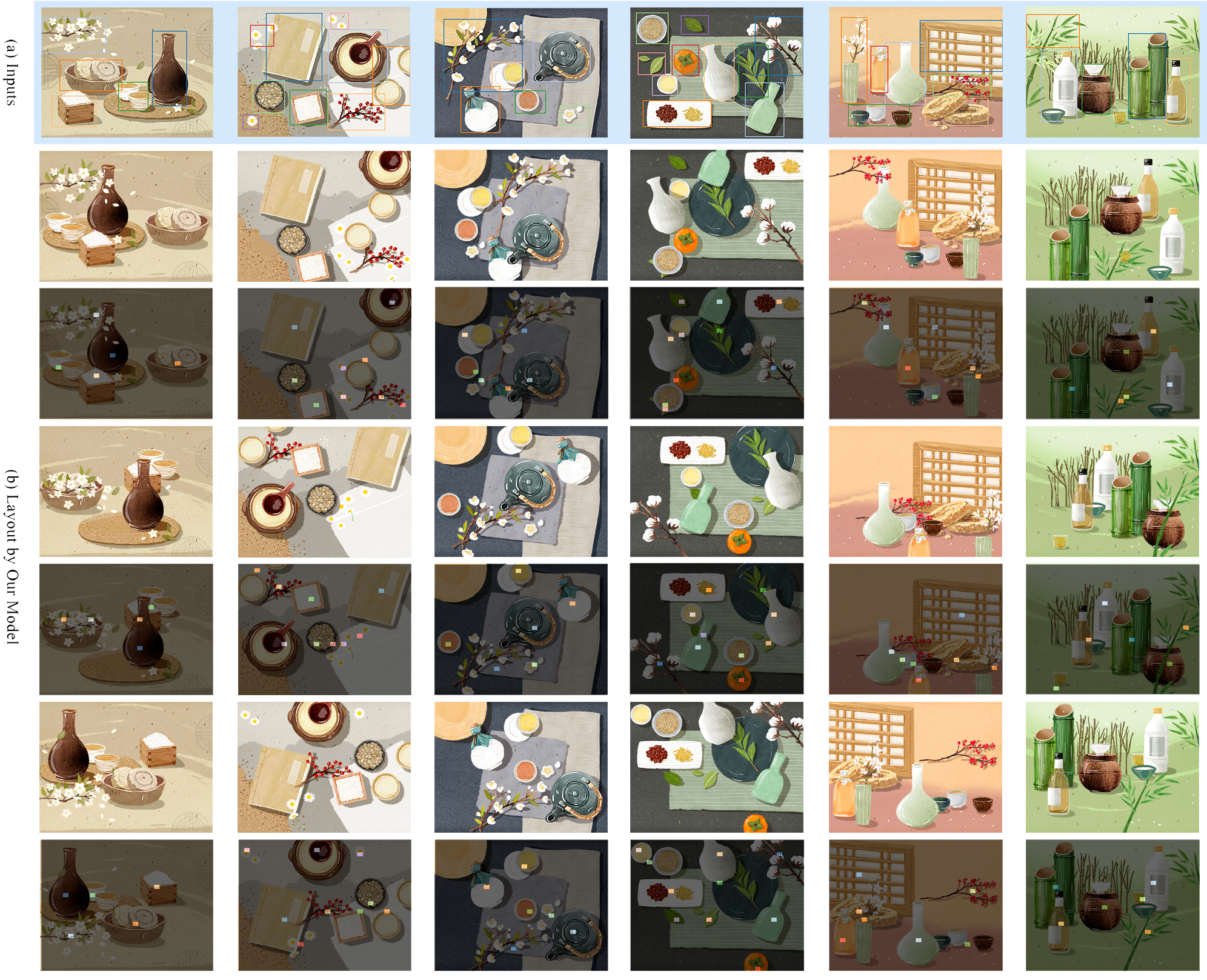}
	\caption{Supplementary materials of Figure~\ref{fig:minsu}.}
	\label{fig:apx3}
\end{figure*}

\begin{figure*}
	\centering
	\includegraphics[width=\linewidth]{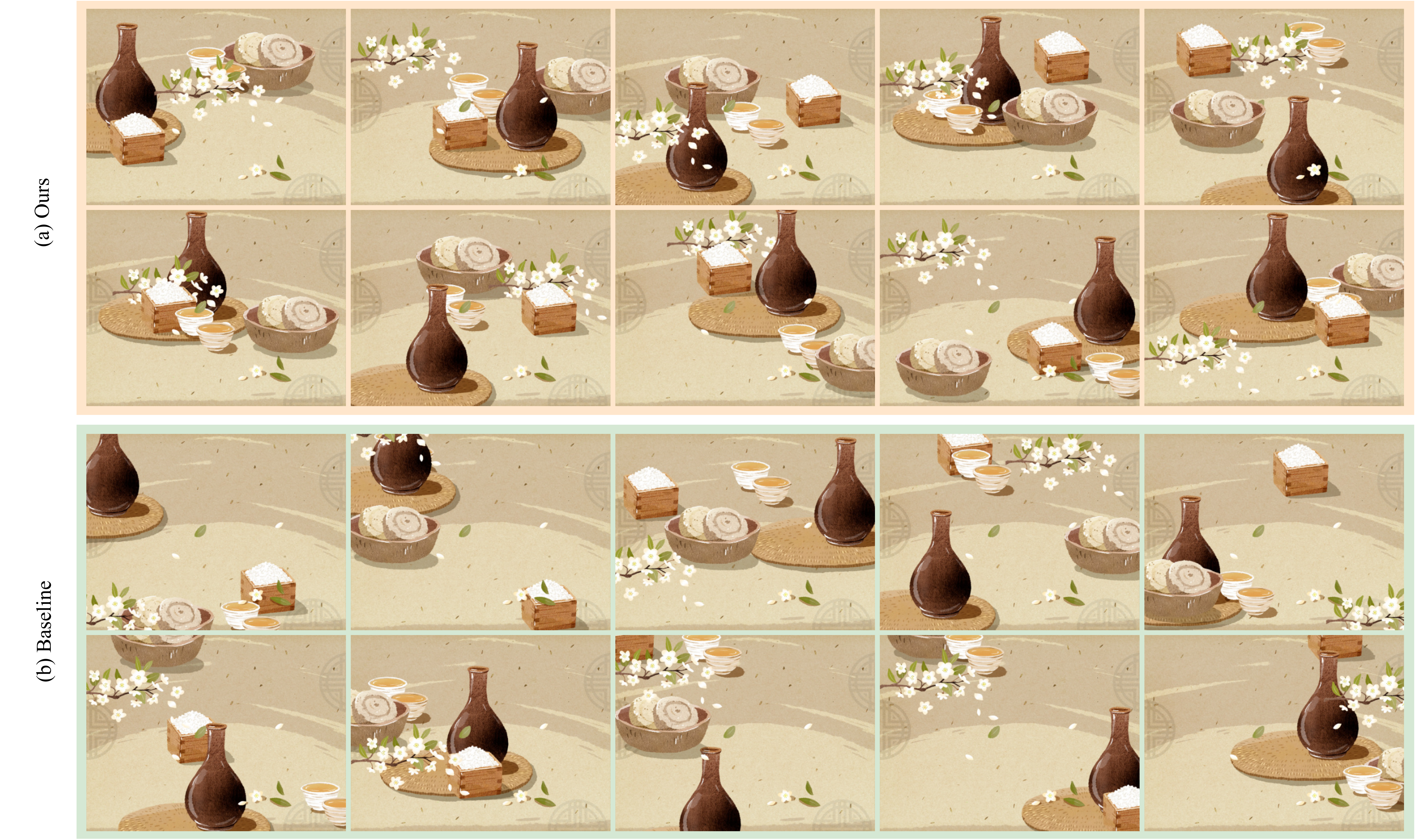}
	\caption{We show ten results from our model and ten results from baseline without selection. The baseline are random layouts generated from uniform noise.}
	\label{fig:10result}
\end{figure*}


\end{document}